\documentclass[11pt]{article}

\usepackage[utf8]{inputenc}
\usepackage[T1]{fontenc}
\usepackage{amsmath,amssymb,amsfonts}
\usepackage{graphicx}
\usepackage{booktabs}
\usepackage{array}
\usepackage{caption}
\usepackage[margin=1in]{geometry}
\usepackage{microtype}
\usepackage{listings}
\usepackage{xurl}
\usepackage[colorlinks=true,linkcolor=blue,citecolor=blue,urlcolor=blue]{hyperref}

\title{\bfseries DCA-MoE: Spatially Adaptive Cross-Layer Fusion and
Density-Routed Experts for Crowd Counting}

\author{Hao Wang\\
BDNRC\\
\texttt{wh1090220084@163.com}}

\date{}

\begin{document}

\maketitle

\begin{abstract}
Crowd counting must recover reliable local density under severe variations in
perspective, head scale, occlusion, and background clutter. Although modern
counting objectives provide strong spatial supervision, many multi-level
decoders still use spatially invariant feature fusion and apply one
receptive-field pattern to every location. We propose DCA-MoE, a framework that
makes both decisions content dependent while retaining a frozen DINOv3 encoder.
Spatially Adaptive Layer Fusion (SALF) predicts position-wise weights over four
aligned backbone features, and Density-Routed Multi-Receptive-Field Experts
(DR-MoE) assigns each location a soft mixture of local, mid-range, and
large-context residual experts. An EBC-style head reconstructs block density,
while DMCount supervision and an auxiliary routing-balance term train the
decoder without updating the backbone. On the NWPU-Crowd validation split, the
strongest paired configuration, based on DINOv3 ViT-L/16, obtains 31.7 MAE and
72.2 RMSE; the matched ViT-B/16 full model obtains a paired 32.2/75.9.
Cross-dataset results remain mixed, and several component baselines currently
report independently selected minima from a single seed. The evidence therefore
supports the feasibility of spatially adaptive fusion and routing, while broader
paired and multi-seed evaluation remains necessary for causal attribution.
\end{abstract}


\section{Introduction}

Crowd counting estimates the number and spatial distribution of people in an
image and supports applications such as public-space monitoring, transportation
management, and safety analysis. The task is difficult because the same image
can contain isolated large heads, heavily occluded small heads, strong
perspective changes, and background patterns that resemble people. Large
benchmarks such as ShanghaiTech \cite{zhang2016single}, UCF-QNRF
\cite{idrees2018composition}, and NWPU-Crowd \cite{wang2021nwpu} expose this
heterogeneity across both scene scale and crowd density.

Deep crowd-counting research has progressed from multi-column scale-specific
regressors \cite{zhang2016single} to dilated single-column networks
\cite{li2018csrnet}, context-aware estimation \cite{liu2019context}, and
point-supervised objectives that reduce dependence on hand-designed Gaussian
density maps \cite{ma2019bayesian,wang2020distribution}. Detection and
localization formulations, including P2PNet \cite{song2021rethinking} and CLTR
\cite{liang2022end}, predict person locations directly, while STEERER
\cite{han2023steerer}, mPrompt \cite{guo2024regressor}, and hybrid
Transformer-CNN designs \cite{zhao2026hybrid} address scale variation, annotation
uncertainty, or long-range context. These approaches improve supervision and
representation, but a persistent architectural question remains: which feature
depth and spatial context should be used at each image location?

Classification-based counting offers a complementary output formulation.
Standard blockwise classification quantizes local counts
\cite{liu2020counting}, and spatial divide-and-conquer adapts the count range
through a hierarchy \cite{xiong2019open}. Vision-language approaches then use
semantic priors from CLIP \cite{radford2021learning}: CrowdCLIP performs weakly
or unsupervised global counting \cite{liang2023crowdclip}, CLIP-Count studies
text-guided zero-shot counting \cite{jiang2023clipcount}, and CLIP-EBC
reconstructs local density through enhanced blockwise classification and
prompt-tuned CLIP features \cite{ma2025clipebc}. These methods improve target
structure or transfer semantic knowledge, but the prediction head alone does not
specify how heterogeneous intermediate visual features should be fused within
one scene.

Multi-scale representation is commonly built through feature pyramids
\cite{lin2017feature}, explicit multi-level fusion \cite{ma2022fusioncount},
multifaceted attention \cite{lin2022boosting}, or dynamic counter mixtures
\cite{wang2023dynamic}. Dynamic convolution further shows that input-conditioned
kernel aggregation can adapt computation to content \cite{chen2020dynamic}.
However, many crowd-counting decoders still select a fixed set of backbone
outputs and combine them through addition, concatenation, or globally shared
attention. Such fusion applies the same feature-depth preference across sparse
and congested regions, even though their localization and context requirements
differ.

Large visual encoders make this issue more consequential. Vision Transformers
\cite{dosovitskiy2021image} expose layer-wise token features, ConvNeXt
\cite{liu2022convnet} provides hierarchical convolutional stages, and
self-supervised systems from DINO \cite{caron2021emerging} and DINOv2
\cite{oquab2024dinov2} to DINOv3 \cite{simeoni2025dinov3} learn transferable
dense representations without task labels. DINOv3 supplies both ViT and ConvNeXt
families with strong intermediate features, but their useful depth is not
uniform: shallow or high-resolution outputs retain local detail, whereas deeper
outputs encode broader context and stronger semantics. Freezing the encoder
reduces trainable cost and preserves the pretrained representation, yet it
places greater responsibility on the decoder to select and transform those
features appropriately.

We introduce DCA-MoE to make two decoder decisions spatially adaptive. SALF
aligns four selected backbone outputs and predicts a distribution over feature
depth at every output location. DR-MoE then predicts a soft spatial distribution
over three residual experts with complementary receptive fields. The router is
supervised indirectly by block-density reconstruction and counting losses; it
does not consume an explicit predicted density map. A balancing objective
discourages persistent expert collapse, following the general principle of
regularized expert routing while retaining dense soft mixtures for spatial
reconstruction.

The contributions are:

\begin{enumerate}
  \item We propose SALF, a position-dependent cross-layer fusion module that
        replaces spatially invariant summation with normalized local
        feature-depth selection.
  \item We propose DR-MoE, a density-supervised spatial router over local,
        mid-range, and large-context experts, together with a balancing term
        that promotes expert utilization.
  \item We integrate both modules with frozen DINOv3 ConvNeXt and ViT encoders,
        an EBC-style density head, and DMCount supervision, and report
        NWPU-Crowd backbone, layer-selection, capacity, and matched single-seed
        component studies. Paired-checkpoint and multi-seed limitations are
        stated explicitly rather than treated as evidence of universal
        superiority.
\end{enumerate}

\section{Related Work}

\subsection{Density regression, point supervision, and localization}

Density-map regression converts point annotations into spatial targets whose
integral gives the image count. MCNN uses parallel columns to cover different
head scales \cite{zhang2016single}, whereas CSRNet replaces explicit columns
with dilated convolutions \cite{li2018csrnet}. Context-aware crowd counting
introduces scale-aware contextual aggregation \cite{liu2019context}. Because
Gaussian target construction can introduce kernel and perspective sensitivity,
Bayesian Loss treats each annotation as a probabilistic contribution
\cite{ma2019bayesian}, and DMCount matches predicted density to point
annotations through optimal transport while preserving global count consistency
\cite{wang2020distribution}. P2PNet directly predicts a set of points
\cite{song2021rethinking}, CLTR models localization with an end-to-end
Transformer \cite{liang2022end}, and STEERER selectively inherits representations
across scales \cite{han2023steerer}. More recent methods use mutual prompting
between regression and segmentation \cite{guo2024regressor} or hybrid
Transformer-CNN U-shaped architectures \cite{zhao2026hybrid}. DCA-MoE does not
replace these supervision paradigms; it uses DMCount-based supervision to study
adaptive feature fusion and spatial receptive-field selection.

\subsection{Classification-based and vision-language counting}

Classification-based methods reduce the long-tailed regression space by
assigning counts to intervals. Standard blockwise classification predicts a class
for each local block \cite{liu2020counting}, while spatial divide-and-conquer
recursively partitions high-count regions \cite{xiong2019open}. EBC refines
blockwise classification through integer-valued bins, noise-aware target
construction, and a joint counting objective \cite{ma2025clipebc}. Separately,
CLIP transfers image-language alignment learned at web scale
\cite{radford2021learning}. CrowdCLIP uses CLIP for weakly or unsupervised
image-level crowd counting \cite{liang2023crowdclip}, and CLIP-Count conditions
zero-shot object counting on text \cite{jiang2023clipcount}. CLIP-EBC combines
CLIP with local density reconstruction \cite{ma2025clipebc}. Our work retains the
structured local EBC target but replaces CLIP prompt tuning with frozen DINOv3
features and focuses on position-wise layer and receptive-field adaptation.

\subsection{Multi-level fusion and scale-aware decoding}

Feature Pyramid Networks established top-down multi-level fusion for dense
prediction \cite{lin2017feature}. Crowd-counting systems similarly combine
features at different depths or scales: FusionCount emphasizes efficient
multi-scale fusion \cite{ma2022fusioncount}, multifaceted attention aggregates
complementary cues \cite{lin2022boosting}, and Dynamic Mixture of Counter Network
selects among counter branches for location-agnostic counting
\cite{wang2023dynamic}. Dynamic convolution generalizes this idea by predicting
input-dependent attention over convolution kernels \cite{chen2020dynamic}. These
studies motivate adaptive processing, but they differ in where adaptation
occurs. SALF predicts feature-depth weights at every spatial position, and
DR-MoE separately predicts position-wise mixtures over receptive-field experts
after fusion. This two-stage factorization distinguishes selecting a
representation level from selecting a spatial operator.

\subsection{Conditional computation and mixture of experts}

Mixture-of-experts models learn a router that assigns inputs to specialized
subnetworks. Sparsely gated MoE layers scale model capacity through conditional
expert activation \cite{shazeer2017outrageously}, Switch Transformers simplify
sparse routing and introduce expert-load regularization \cite{fedus2022switch},
and V-MoE extends sparse expert routing to vision Transformers
\cite{riquelme2021scaling}. Sparse MoE primarily targets computational scaling at
token level. DCA-MoE instead uses dense soft routing on a shared spatial grid:
all three lightweight residual experts produce features, and the router blends
them per location for density reconstruction. The balancing loss is inspired by
load-regularized MoE, but the objective here is stable spatial specialization
rather than reduced floating-point cost.

\subsection{Self-supervised visual foundation features}

ViT represents images as token sequences \cite{dosovitskiy2021image}, while
ConvNeXt modernizes hierarchical convolutional networks using design choices
associated with Transformers \cite{liu2022convnet}. DINO demonstrates that
self-distillation can produce semantically meaningful ViT attention and
transferable features \cite{caron2021emerging}, and DINOv2 improves the
robustness of general-purpose visual representations \cite{oquab2024dinov2}.
DINOv3 further scales self-supervised training and provides dense features across
ViT and ConvNeXt model families \cite{simeoni2025dinov3}. Unlike methods that
fine-tune an entire task backbone, we freeze DINOv3 and train only the alignment,
fusion, expert-routing, and density-prediction modules. This setting isolates
decoder adaptation and reduces trainable parameters, but it does not establish
that DINOv3 is intrinsically superior to other pretrained encoders without a
matched backbone study.

\section{Method}

\subsection{Overview}

Given an image $I$, a frozen DINOv3 encoder extracts four intermediate
representations:
\[
\{X_i\}_{i=1}^{4}=f_{\mathrm{DINOv3}}(I).
\]
Each representation is projected to $C$ channels and resized to
$H/8\times W/8$, yielding $F_i$. SALF predicts a position-dependent distribution
over the four features. The fused representation is smoothed and passed to
DR-MoE, which mixes three receptive-field experts at every spatial location. An
EBC-style classifier produces local count-bin probabilities from which the
density map and image count are reconstructed.

\begin{figure}[t]
  \centering
  \includegraphics[width=\linewidth]{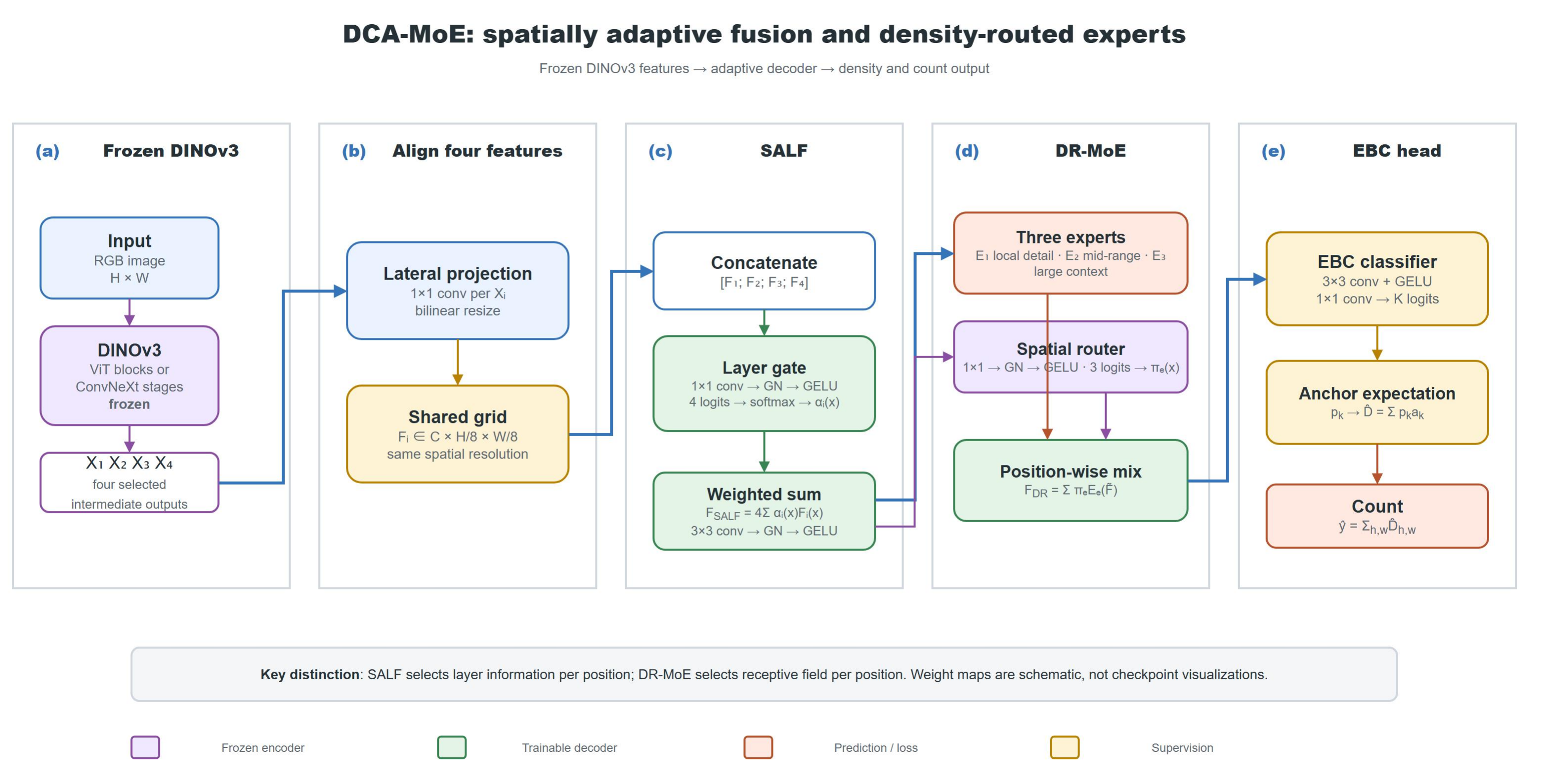}
  \caption{Overall DCA-MoE architecture. Four frozen DINOv3 outputs are aligned
  to a shared $1/8$ grid, fused by position-wise SALF weights, processed by
  three receptive-field experts under a spatial router, and converted to density
  through EBC anchor expectation. The diagram uses one-to-one stage connections
  to keep the feature path readable; the editable source is provided in
  \texttt{figures/drawio/fig1\_dca\_moe\_architecture\_drawio.drawio}.}
  \label{fig:arch}
\end{figure}

The DINOv3 parameters are frozen throughout training. The lateral projections,
SALF, smoothing block, DR-MoE, and prediction head are trainable.

\subsection{Multi-level feature extraction and alignment}

For a ConvNeXt backbone, the adapter uses all four stages with output strides 4,
8, 16, and 32. For ConvNeXt-Base, their channel dimensions are 128, 256, 512,
and 1024. For a ViT backbone, four Transformer blocks are selected. The default
rule uses the last four blocks: $[8,9,10,11]$ for 12 blocks, $[20,21,22,23]$ for
24 blocks, and $[28,29,30,31]$ for 32 blocks. Alternative block sets are exposed
through \texttt{--vit\_layers} for controlled layer-selection experiments.

The aligned feature at level $i$ is
\[
F_i=\operatorname{Resize}_{H/8,W/8}\left(W_i * X_i+b_i\right),
\qquad F_i\in\mathbb{R}^{C\times H/8\times W/8},
\]
where $W_i$ is a learnable $1\times1$ projection and bilinear interpolation
performs spatial alignment.

\subsection{Spatially Adaptive Layer Fusion}

SALF concatenates the four aligned features and predicts four logits at every
spatial location:
\[
A=g_{\mathrm{SALF}}([F_1;F_2;F_3;F_4]),
\qquad
\alpha_i(x)=\frac{\exp(A_i(x))}{\sum_{j=1}^{4}\exp(A_j(x))}.
\]
The cross-layer fused representation is
\[
F_{\mathrm{SALF}}(x)=4\sum_{i=1}^{4}\alpha_i(x)F_i(x).
\]
The factor 4 preserves the magnitude of fixed four-feature summation when
$\alpha_i(x)=1/4$. The final SALF convolution is initialized to zero, so training
starts from this uniform and stable state before learning spatial
specialization. A $3\times3$ convolution, GroupNorm, and GELU then produce the
smoothed representation $\widetilde{F}$.

\subsection{Density-Routed Multi-Receptive-Field Experts}

DR-MoE contains three residual experts designed for complementary context
ranges:

\begin{table}[t]
  \centering
  \caption{Density-routed multi-receptive-field expert design.}
  \label{tab:experts}
  \begin{tabular}{lll}
    \toprule
    Expert & Main spatial operator & Intended role \\
    \midrule
    $E_1$ & standard $3\times3$ convolution & local detail and small-head localization \\
    $E_2$ & depthwise $3\times3$ convolution, dilation 2 & mid-range context \\
    $E_3$ & depthwise $7\times7$ convolution & large-context and dense-region reasoning \\
    \bottomrule
  \end{tabular}
\end{table}

Each expert includes a following $1\times1$ projection, GroupNorm, GELU, and
residual connection. The router predicts a three-way spatial distribution from
$\widetilde{F}$:
\[
\pi_e(x)=\operatorname{Softmax}_e\big(g_{\mathrm{route}}(\widetilde{F})(x)\big),
\qquad
F_{\mathrm{DR}}(x)=\sum_{e=1}^{3}\pi_e(x)\,E_e(\widetilde{F})(x).
\]
The routing is dense and soft: all expert outputs are computed, and $\pi_e(x)$
controls their local contribution. The router receives fused visual features and
is optimized through the downstream density and counting losses. Thus, density
awareness is learned implicitly from supervision rather than supplied by an
explicit density-map branch.

\begin{figure}[t]
  \centering
  \includegraphics[width=\linewidth]{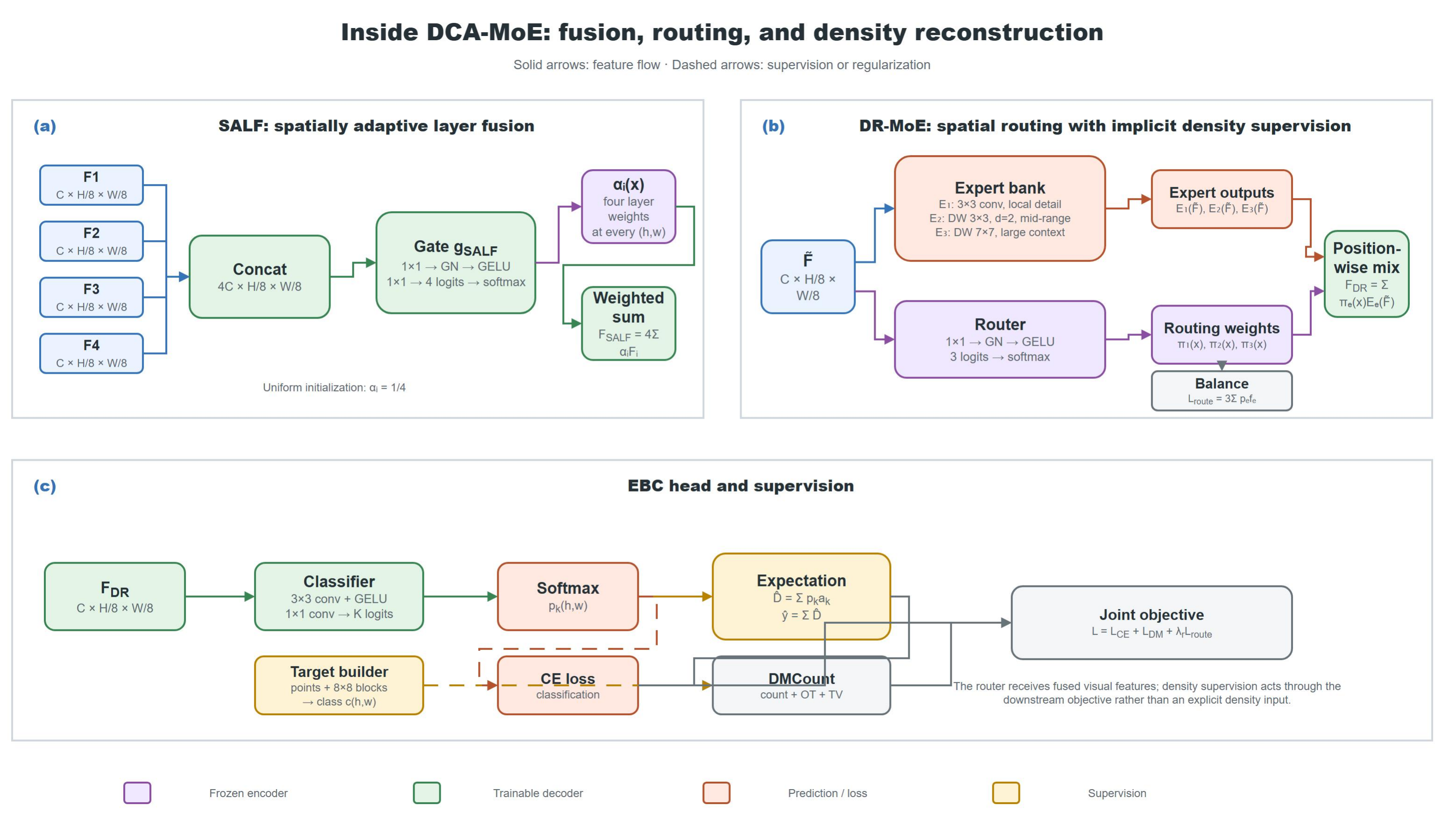}
  \caption{Detailed SALF, DR-MoE, and EBC computation. Solid arrows show feature
  flow, while dashed arrows show supervision or regularization. The expert bank
  lists the three receptive-field operators, and the router produces
  position-wise weights that enter the final mixture through a separate path.
  The diagram also identifies tensor dimensions, anchor-based density
  reconstruction, and the CE/DMCount supervision paths. The editable source is
  provided in
  \texttt{figures/drawio/fig2\_dca\_moe\_detail\_drawio.drawio}.}
  \label{fig:detail}
\end{figure}

To discourage routing collapse, let $p_e$ be the mean routing probability and
$f_e$ the fraction of locations whose maximum-probability expert is $e$. The
balancing loss is
\[
\mathcal{L}_{\mathrm{route}}=3\sum_{e=1}^{3}p_e f_e.
\]

\subsection{Block classification and density reconstruction}

At output reduction 8, the NWPU local-density target uses five intervals:
\[
\mathcal{B}=\{[0,0],[1,1],[2,2],[3,3],[4,+\infty)\},
\qquad
\mathbf{a}=[0,1,2,3,4.21931].
\]
Let $z_{b,h,w,k}$ be the class logit and $p_{b,h,w,k}$ its softmax probability.
The predicted density and image count are
\[
\widehat{D}_{b,h,w}=\sum_{k=1}^{5}p_{b,h,w,k}\,a_k,
\qquad
\widehat{y}_b=\sum_{h,w}\widehat{D}_{b,h,w}.
\]

\subsection{Training objective}

The block-classification term is the sum of the pixel-wise cross-entropy values
over the output grid:
\[
\mathcal{L}_{\mathrm{CE}}=
\frac{1}{B}\sum_{b=1}^{B}\sum_{h,w}
\operatorname{CE}(z_{b,h,w},c_{b,h,w}).
\]
The density-counting component follows DMCount:
\[
\mathcal{L}_{\mathrm{DM}}=
\mathcal{L}_{\mathrm{count}}
 + \lambda_{\mathrm{OT}}\mathcal{L}_{\mathrm{OT}}
 + \lambda_{\mathrm{TV}}\mathcal{L}_{\mathrm{TV}}.
\]
The complete objective is
\[
\mathcal{L}_{\mathrm{total}}=
\mathcal{L}_{\mathrm{CE}}+
\lambda_{\mathrm{count}}\mathcal{L}_{\mathrm{DM}}+
\lambda_{\mathrm{route}}\mathcal{L}_{\mathrm{route}},
\]
where $\lambda_{\mathrm{count}}=1$, $\lambda_{\mathrm{OT}}=0.1$, and
$\lambda_{\mathrm{TV}}=0.01$. The reported DCA-MoE runs use
$\lambda_{\mathrm{route}}=0.1$; a matched no-balance experiment remains part of
the required ablation.

\section{Experimental Results}

\subsection{Experimental Setup}

\subsubsection{Dataset and Preprocessing}

The completed experimental stage uses NWPU-Crowd. Its official split contains
3,109 training images, 500 validation images, and 1,500 test images. The
training and validation annotations are read from the official point labels. The
test split is used only for prediction because its public annotations are not
available in the ordinary evaluation workflow. The cross-dataset stage extends
the same ViT-B/16 architecture to ShanghaiTech Part A (SHA), ShanghaiTech Part B
(SHB), and UCF-QNRF. Their official train/test splits contain 300/182, 400/316,
and 1,201/334 images, respectively.

For compatibility with the training code, the official SHA/SHB
\texttt{test\_data} directories and the UCF-QNRF \texttt{Test} directory are
stored locally under the split name \texttt{val}. This naming is an
implementation detail: these images remain official test data and must not be
used for hyperparameter selection or repeated best-checkpoint selection. We use
a deterministic internal holdout from each official training set to select the
training duration and checkpoint rule. After fixing these choices, the model is
retrained on the complete official training set and evaluated once on the
official test set. The split indices, selected duration, and random seeds must be
retained with the final results.

The preprocessing pipeline preserves the image aspect ratio, scales point
coordinates with the image, and adjusts image sides to multiples of 32. The
default minimum image side is 448 pixels and the NWPU maximum side is 3072
pixels. The processed layout is:

\begin{lstlisting}
data/nwpu/train/images/*.jpg
data/nwpu/train/labels/*.npy
data/nwpu/val/images/*.jpg
data/nwpu/val/labels/*.npy
data/nwpu/test/images/*.jpg
data/{sha,shb,qnrf}/{train,val}/{images,labels}/*
\end{lstlisting}

\subsubsection{Data Augmentation}

The experiments use the following point-consistent augmentation pipeline:

\begin{itemize}
  \item random resized crop to the configured input size with scale range
        $[0.5, 4.0]$;
  \item random horizontal flip.
  \item independent color jitter with brightness, contrast, and saturation
        strength 0.1 and probability 0.2;
  \item $5\times5$ Gaussian blur with sigma range $[0.1, 5.0]$ and probability
        0.2;
  \item pepper-and-salt noise with saltiness and spiciness $1\times10^{-3}$ and
        probability 0.5.
\end{itemize}

Geometric operations update point coordinates. Validation uses the unaugmented
image and point annotations.

The augmentation operations are applied before ImageNet normalization. The
primary experiment uses 448 pixels, while the controlled resolution experiment
B7-224 uses a $224\times224$ training crop. For B7-224, the validation sliding
window and stride are also set to 224 pixels.

\subsubsection{Optimization and Implementation Details}

The primary NWPU experiments use the following settings:

\begin{table}[t]
  \centering
  \caption{Primary NWPU-Crowd training and evaluation settings.}
  \label{tab:settings}
  \small
  \begin{tabular}{@{}p{0.24\textwidth}p{0.70\textwidth}@{}}
    \toprule
    Item & Setting \\
    \midrule
    Input size & $448\times448$ training crop; $224\times224$ for B7-224 \\
    Output reduction & 8 \\
    Backbone & frozen DINOv3 ConvNeXt-Base or ViT variant \\
    Decoder channels & 128 for the legacy fixed-fusion controls; 192 for ViT-S/16 and ViT-S+/16; 384 for B7-MoE, ConvNeXt-Small, and ConvNeXt-Tiny DCA-MoE; 512 for ConvNeXt-Base DCA-MoE, ConvNeXt-Large width-control, and ViT-L/16; 768 for ConvNeXt-Large DCA-MoE; 1024 for H+-MoE \\
    Optimizer & AdamW \\
    Initial learning rate & $1\times10^{-4}$ \\
    Weight decay & $1\times10^{-4}$ \\
    Warm-up & 5 epochs \\
    Scheduler & cosine decay to 0.01 of the initial rate \\
    Mixed precision & BF16 AMP when supported \\
    Batch size & report per-GPU and global batch size separately \\
    Random seed & 42 \\
    Primary count loss & DMCount \\
    Validation & from \texttt{eval\_start}, then every \texttt{eval\_freq} epochs \\
    Checkpoint & best validation MAE, best validation RMSE, and last epoch \\
    \bottomrule
  \end{tabular}
\end{table}

The DINOv3 backbone is evaluated in \texttt{eval()} mode and has
\texttt{requires\_grad=False}. Only the adapter and density head are updated.

The reported B6, B7, and B7-224 results in this draft were produced with the
original 128-channel decoder. Their reproduction commands therefore set
\texttt{--fpn\_channels 128} explicitly. The model-size-aware widths are a
separate capacity-scaling configuration and require new training runs.

For comparison, the CLIP-EBC paper reports Adam with an initial learning rate of
$1\times10^{-4}$, cosine annealing, batch size 8, and visual prompt tuning for
ViT models. These settings differ from the current DINOv3 implementation, which
uses AdamW, five warm-up epochs, a minimum learning-rate ratio of 0.01, and a
user-specified per-GPU batch size. In an eight-GPU run, \texttt{--batch\_size 8}
means a global batch size of 64 in the current script, not the paper's batch size
of 8.

The cross-dataset study consequently uses two comparison levels. First, the main
benchmark table quotes the CLIP-EBC values published in \cite{ma2025clipebc} and
labels them as external reference results. Second, the matched DINOv3 A0/A4 study
keeps the backbone, selected layers $[8,9,10,11]$, FPN384 width, data processing,
optimizer, global batch size, training duration, checkpoint rule, and seed fixed
while changing only the decoder. If CLIP-EBC is re-run, its official model and
training protocol are preserved instead of forcing it into the DINOv3
optimization recipe; those measurements are reported in separate
\texttt{reproduced} rows with the repository revision and full command. This
separation avoids conflating method comparison with optimizer or pretraining
differences.

\subsubsection{Evaluation Metrics}

\begin{figure}[t]
  \centering
  \includegraphics[width=\linewidth]{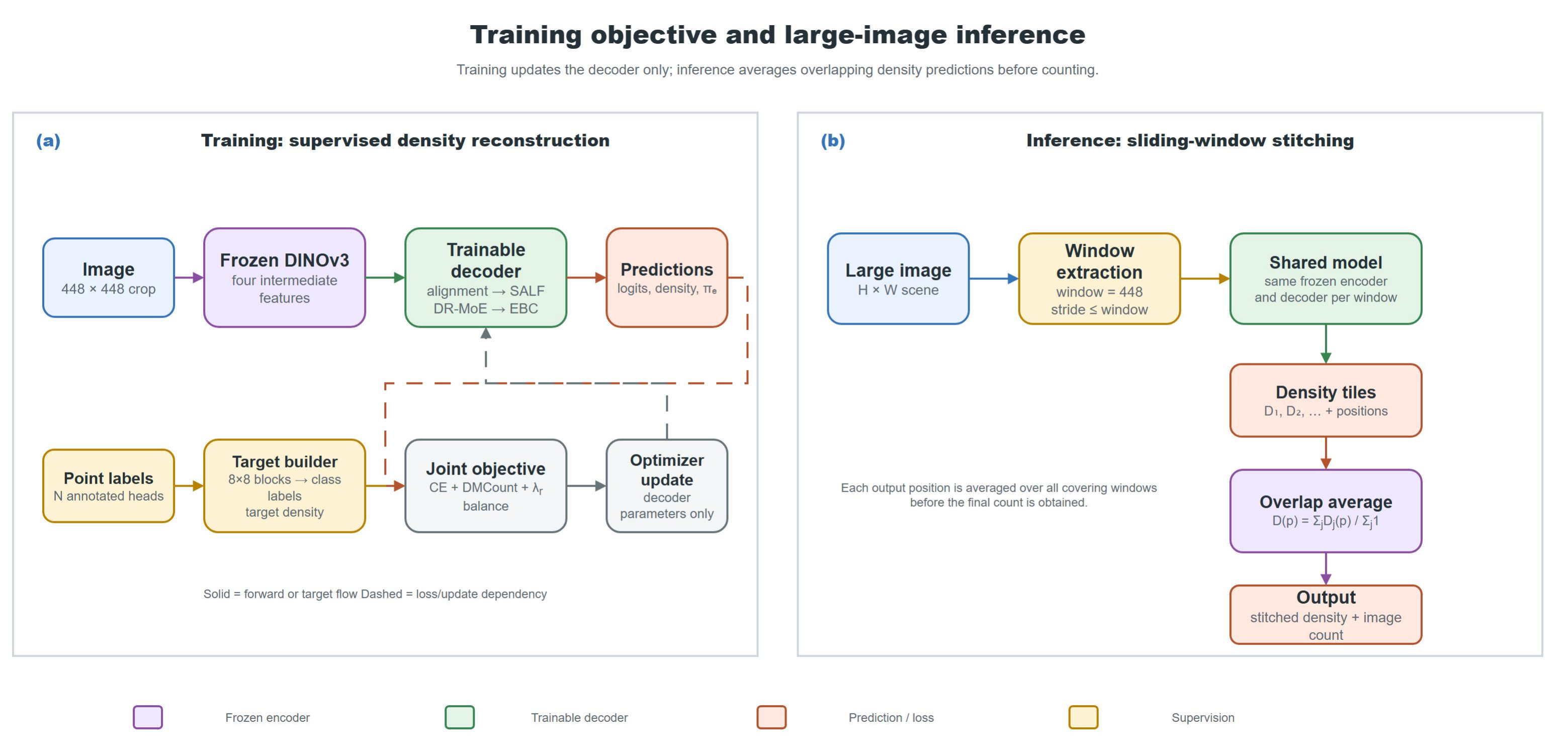}
  \caption{Training and large-image inference. In the training panel, solid
  arrows denote forward or target construction flow and dashed arrows denote
  loss or optimizer-update dependencies. In the inference panel, a shared model
  predicts density tiles for each window, and overlapping predictions are
  averaged at every covered output position before the image count is obtained.
  The editable source is provided in
  \texttt{figures/drawio/fig3\_dca\_moe\_training\_inference\_drawio.drawio}.}
  \label{fig:inference}
\end{figure}

For $N$ images with predicted counts $\hat{y}_i$ and ground-truth counts $y_i$,
we report:
\[
\operatorname{MAE}=\frac{1}{N}\sum_{i=1}^{N}|\hat{y}_i-y_i|,
\]
\[
\operatorname{RMSE}=\sqrt{\frac{1}{N}\sum_{i=1}^{N}(\hat{y}_i-y_i)^2}.
\]
MAE is used as the primary checkpoint-selection criterion. RMSE is reported
because it exposes sensitivity to severe counting errors on highly crowded
images.

\subsection{Main Benchmark Results}

\subsubsection{Cross-Dataset Comparison with CLIP-EBC}

Table~\ref{tab:cross} compares DINOv3-DCA-MoE with the published CLIP-EBC results
from Table~III of \cite{ma2025clipebc}. SHA, SHB, and UCF-QNRF are evaluated on
their official test sets, whereas NWPU is evaluated on its labeled validation
split. All entries are reported as MAE/RMSE, lower values are better, and the
displayed values are rounded to one decimal place.

\begin{table}[t]
  \centering
  \caption{Cross-dataset comparison with CLIP-EBC (MAE/RMSE).}
  \label{tab:cross}
  \resizebox{\textwidth}{!}{%
  \begin{tabular}{llrrrr}
    \toprule
    Method & Backbone and adaptation & SHA Test MAE/RMSE & SHB Test MAE/RMSE & UCF-QNRF Test MAE/RMSE & NWPU Val MAE/RMSE \\
    \midrule
    CLIP-EBC & CLIP ResNet50 & 54.0/83.2 & 6.0/10.1 & 80.5/136.6 & 38.6/90.3 \\
    CLIP-EBC & CLIP ViT-B/16 & 52.5/85.9 & 6.6/10.5 & 80.3/139.3 & 36.6/81.7 \\
    DINOv3-DCA-MoE & DINOv3 ViT-B/16 & 55.4/95.0 & 8.1/13.3 & 77.7/141.7 & 32.2/75.9 \\
    \bottomrule
  \end{tabular}}
\end{table}

CLIP-EBC ViT-B/16 is the primary external reference because it uses the same
ViT-B/16 architecture class. On SHA, DINOv3-DCA-MoE obtains 55.4/95.0, compared
with 52.5/85.9 for CLIP-EBC ViT-B/16; its MAE and RMSE are therefore higher by
2.9 and 9.1, respectively. On SHB, DINOv3-DCA-MoE reports 8.1/13.3 versus
6.6/10.5, corresponding to increases of 1.5 MAE and 2.8 RMSE. Thus, the
DINOv3-based model is weaker than the ViT-B/16 CLIP-EBC reference on both
ShanghaiTech subsets.

The comparison is mixed on UCF-QNRF. DINOv3-DCA-MoE reduces MAE from 80.3 to
77.7, an absolute improvement of 2.6, but increases RMSE from 139.3 to 141.7, a
deterioration of 2.4. On NWPU validation, it improves both metrics, reducing MAE
from 36.6 to 32.2 and RMSE from 81.7 to 75.9, corresponding to absolute
improvements of 4.4 and 5.8, respectively.

Overall, Table~\ref{tab:cross} does not show a uniform cross-dataset advantage:
DINOv3-DCA-MoE performs best relative to CLIP-EBC on NWPU, exhibits an
MAE-RMSE trade-off on UCF-QNRF, and remains weaker on SHA and SHB. Moreover,
CLIP-EBC uses CLIP pretraining and visual prompt tuning, whereas DCA-MoE uses a
frozen DINOv3 encoder and a different decoder. The table therefore measures
benchmark-level competitiveness rather than providing a controlled attribution of
gains to SALF or DR-MoE; component attribution is instead based on the matched
A0-A4 comparison in Table~\ref{tab:ablation}.

\subsubsection{Detailed NWPU-Crowd Validation Results}

Table~\ref{tab:nwpu_val} consolidates published results that are explicitly
reported on the labeled NWPU-Crowd validation split and the best result retained
for each exact DINOv3 backbone in our experiments. All entries are reported as
MAE/RMSE, lower values are better, and the displayed values are rounded to one
decimal place. Results from the NWPU test split, cross-domain evaluation without
target adaptation, and papers whose original tables assign the quoted values to
another dataset are excluded. The published rows are taken from CLIP-EBC
Table~III \cite{ma2025clipebc}, mPrompt Table~1 \cite{guo2024regressor}, and
HMSTUNet Table~2 \cite{zhao2026hybrid}.

\begin{table}[t]
  \centering
  \caption{Detailed NWPU-Crowd validation comparison (MAE/RMSE).}
  \label{tab:nwpu_val}
  \small
  \begin{tabular}{llr}
    \toprule
    Method & Backbone or reported setting & NWPU Val MAE/RMSE \\
    \midrule
    P2PNet & paper-reported setting & 77.4/362.0 \\
    DMCount & paper-reported setting & 70.5/357.6 \\
    CLTR & paper-reported setting & 61.9/246.3 \\
    mPrompt & HRNet-W40-C & 50.2/219.0 \\
    HMSTUNet & ConvNeXt-Small & 43.2/119.6 \\
    CLIP-EBC & CLIP ResNet50 & 38.6/90.3 \\
    CLIP-EBC & CLIP ViT-B/16 & 36.6/81.7 \\
    DINOv3-DCA-MoE & DINOv3 ConvNeXt-Tiny & 36.5/94.0 \\
    DINOv3-DCA-MoE & DINOv3 ConvNeXt-Small & 40.3/114.7 \\
    DINOv3-DCA-MoE & DINOv3 ConvNeXt-Base & 38.3/122.6 \\
    DINOv3-DCA-MoE & DINOv3 ConvNeXt-Large & 39.9/124.0 \\
    DINOv3-DCA-MoE & DINOv3 ViT-S/16 & 39.3/104.4* \\
    DINOv3-DCA-MoE & DINOv3 ViT-S+/16 & 36.5/87.8 \\
    DINOv3-DCA-MoE & DINOv3 ViT-B/16 & 32.2/75.9 \\
    DINOv3-DCA-MoE & DINOv3 ViT-L/16 & 31.7/72.2 \\
    DINOv3-DCA-MoE & DINOv3 ViT-H+/16 & 31.0/72.7* \\
    \bottomrule
  \end{tabular}

  \vspace{4pt}
  {\footnotesize *The marked MAE and RMSE values are independently selected
  minima and do not represent one checkpoint. Unmarked DINOv3 rows report both
  metrics at the same checkpoint. Each exact DINOv3 backbone appears once; when
  several runs used that backbone, Table~\ref{tab:nwpu_val} retains its best
  available result under this reporting rule.}
\end{table}

For the architecture-matched ViT-B/16 comparison, DINOv3-DCA-MoE improves on
CLIP-EBC by 4.4 MAE and 5.8 RMSE. Among our clean paired results, ViT-L/16
provides the lowest RMSE at 31.7/72.2. These published-reference comparisons are
not controlled experiments because the methods use different pretraining,
adaptation, and optimization protocols.

Backbone scaling is non-monotonic. In particular, DINOv3 ConvNeXt-Tiny
outperforms the Small, Base, and Large variants on both metrics, while ViT-L/16
improves over ViT-B/16 and the independently selected ViT-H+/16 minima do not
establish a better paired operating point. These comparisons also change decoder
width, feature-depth selection, training details, or checkpoint-selection rules
across rows. Published methods additionally use different pretraining and
adaptation protocols. Table~\ref{tab:nwpu_val} consequently measures
benchmark-level competitiveness rather than serving as a controlled attribution
of performance differences to the backbone or DCA-MoE components.

\subsection{Qualitative Positive and Negative Cases}

\begin{figure}[t]
  \centering
  \includegraphics[width=\linewidth]{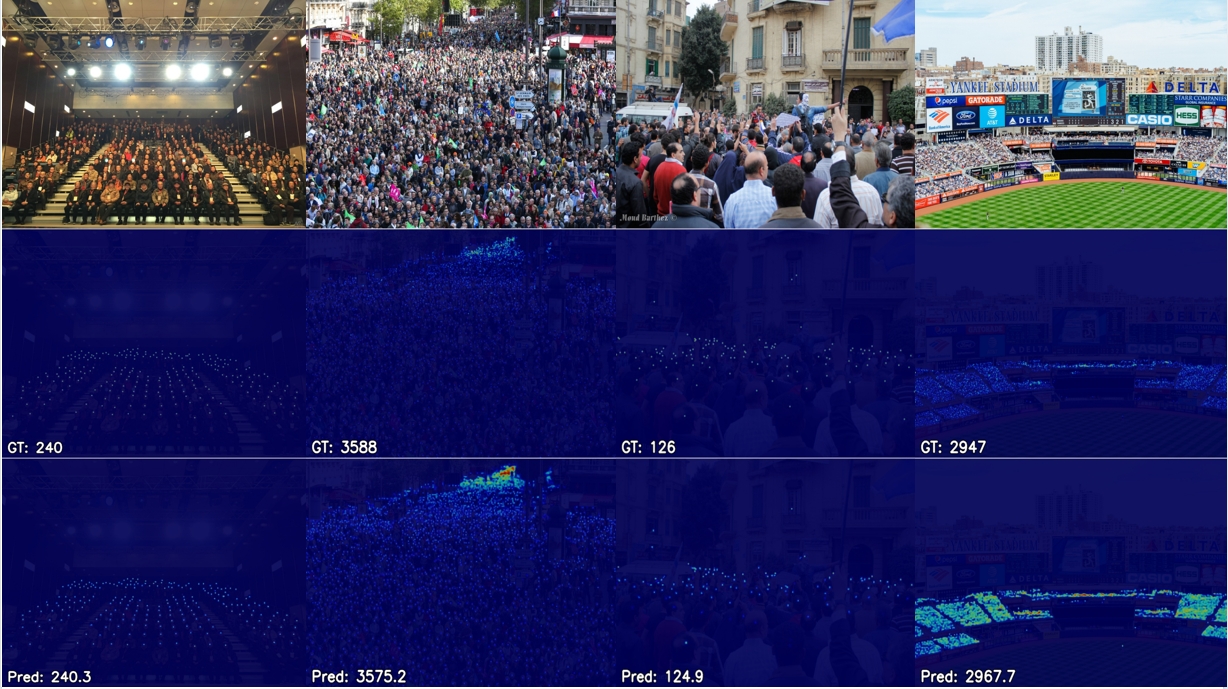}
  \caption{Qualitative comparison on positive samples. From left to right, the
  GT counts are 240, 3588, 126, and 2947; the predicted counts are 240.3,
  3575.2, 124.9, and 2967.7, respectively. Each column contains the input,
  GT density map, and predicted density map from top to bottom.}
  \label{fig:positive_cases}
\end{figure}

\begin{figure}[t]
  \centering
  \includegraphics[width=\linewidth]{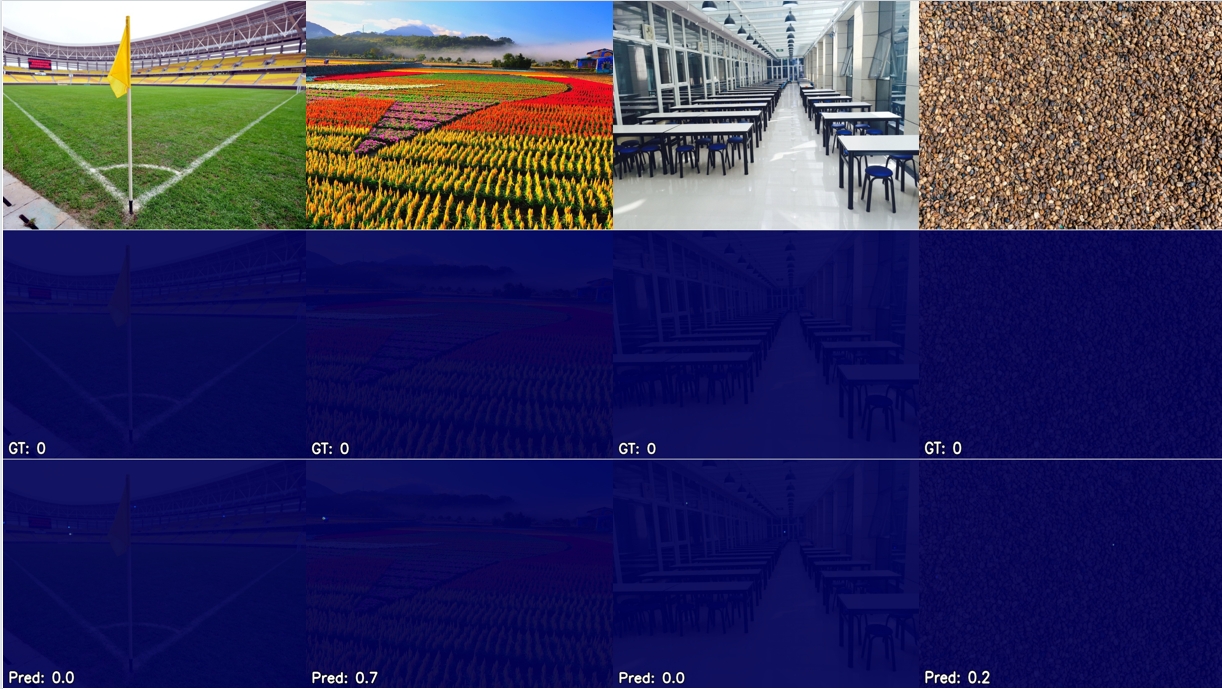}
  \caption{Qualitative comparison on negative samples. The four inputs contain
  no annotated people, so every GT count is zero. From left to right, the
  predicted counts are 0.0, 0.7, 0.0, and 0.2. Each column contains the input,
  GT density map, and predicted density map from top to bottom.}
  \label{fig:negative_cases}
\end{figure}

Figures~\ref{fig:positive_cases} and~\ref{fig:negative_cases} visualize the
estimated density for representative positive and negative images. In each
column, the input image is shown in the first row, the ground-truth (GT) density
map in the second row, and the predicted density map in the third row. The
counts printed in the density-map rows are obtained by summing the corresponding
maps.

For the positive cases in Figure~\ref{fig:positive_cases}, the predicted maps
place density in the same image regions as the GT maps across a seated indoor
audience, an extremely dense street scene, a moderate-density outdoor crowd,
and a large stadium. The absolute count errors are 0.3, 12.8, 1.1, and 20.7,
respectively. The two high-count examples retain their broad crowd support,
while the sparse and moderate-density examples preserve localized responses.
These cases are qualitative illustrations only; they do not replace the
split-level MAE and RMSE reported above.

Figure~\ref{fig:negative_cases} probes false positive responses to a stadium
corner, repeated flower rows, an empty classroom, and gravel texture. The model
suppresses density almost completely on the stadium and classroom images.
Although weak isolated activations remain for the flower field and gravel, their
integrated counts are only 0.7 and 0.2. This behavior indicates that the model
does not systematically convert repeated non-human structure or fine texture
into substantial crowd density on these examples.

\subsection{Component Ablation}

The existing backbone-scale and decoder-width runs are exploratory
configurations, not a strict component ablation. The following matrix defines the
experiment required to support the paper's central claim. All rows must use the
same DINOv3 ViT-B/16 checkpoint, blocks $[8,9,10,11]$,
\texttt{fpn\_channels=384}, input size, augmentation, DMCount weights, global
batch size, optimizer schedule, validation schedule, and random seed. The only
changed factors are SALF, DR-MoE, and the balancing loss. Metric values in the
matrix are rounded to one decimal place.

\begin{table}[t]
  \centering
  \caption{Matched ViT-B/16 component ablation on NWPU-Crowd (MAE and RMSE).}
  \label{tab:ablation}
  \begin{tabular}{llllrr}
    \toprule
    Variant & Layer fusion & Expert block & Balance loss & MAE & RMSE \\
    \midrule
    A0 & fixed sum & shared $3\times3$ decoder & off & 36.2 & 84.7 \\
    A1 & SALF & shared $3\times3$ decoder & off & 36.0 & 87.6 \\
    A2 & fixed sum & DR-MoE & on & 36.1 & 89.1 \\
    A3 & SALF & DR-MoE & off & 32.8 & 77.6 \\
    A4 (DCA-MoE) & SALF & DR-MoE & on & 32.2 & 75.9 \\
    \bottomrule
  \end{tabular}
\end{table}

The implementation now supports the A1 SALF-only path through
\texttt{--decoder\_variant salf\_only}. A1 retains the matched ViT-B/16 backbone,
default last-four blocks $[8,9,10,11]$, FPN384 decoder width, single shared
decoder, DMCount objective, and validation protocol of A0/A4. It creates the
spatial cross-layer fusion gate but omits both the three experts and the density
router; consequently, no router balance term is produced. A1 reaches a best
validation MAE of 35.9776 at epoch 1461 and a best validation RMSE of 87.6369 at
epoch 1565. At epoch 2000, it reports MAE 39.7481 and RMSE 122.3597. The MAE and
RMSE minima are independently selected and do not represent one paired
checkpoint. The epoch-2000 training log reports total loss 98.306721, CE loss
94.687319, TV loss 52.676000, count loss 3.092643, and OT loss approximately
$-0.000001$.

The implementation also supports the A2 MoE-only path through
\texttt{--decoder\_variant moe\_only}. A2 retains the same matched protocol but
uses fixed cross-layer summation followed by the three experts and density
router, with \texttt{router\_loss\_weight=0.1}; it does not create the SALF gate.
The single-seed controlled run reaches a best validation MAE of 36.1265 at epoch
1478 and a best validation RMSE of 89.0765 at epoch 1299. These minima are
independently selected and do not represent one paired checkpoint. At epoch 2000,
it reports MAE 39.2971 and RMSE 116.4205. The epoch-2000 training log reports
total loss 104.556428, CE loss 100.684504, TV loss 56.443647, count loss
3.203058, OT loss approximately $-0.000001$, and router balance loss 1.044292.
The command used \texttt{--decoder\_variant moe\_only}, but its output root was
named \texttt{A1\_salf\_only\_seed42}; the checkpoint path below therefore
preserves the actual directory name while the run is classified as A2 by the
decoder variant.

The A3 Full w/o balance run uses the same SALF, three receptive-field experts,
spatial router, ViT-B/16 backbone, last-four layers, and FPN384 setting as A4,
but sets \texttt{router\_loss\_weight=0}. The seed-42 run reaches a best
validation MAE of 32.7986 at epoch 1354 and a best validation RMSE of 77.6230 at
epoch 1377. These minima are independently selected and do not form one paired
checkpoint. At epoch 2000, A3 reports MAE 35.4477 and RMSE 91.5442; the training
log reports total loss 97.697590, CE loss 94.388890, TV loss 52.502437, count
loss 2.783675, and OT loss approximately $-0.000001$. Because the balance weight
is zero, no balance term is added to the training objective even though the
router remains active.

The matched A4 Full DCA-MoE result is 32.2124 MAE and 75.8733 RMSE, with both
best values reached at epoch 1293. A4 uses SALF, three receptive-field experts,
the spatial density router, and the router balancing loss under the same
ViT-B/16, last-four-layer, FPN384 protocol. Compared with A3's independently
selected minima, A4 reduces MAE by 1.79\% and RMSE by 2.25\%. This seed-42
comparison supports a modest benefit from router balancing, but paired-checkpoint
metrics and multi-seed statistics remain required.

For each row, MAE and RMSE must ultimately be reported at the same selected
checkpoint, with the checkpoint selected by validation MAE. A0, A1, A2, and A3
currently list independently selected minima because their best MAE and best RMSE
occur at different epochs; A4 is already paired at epoch 1293. A three-seed mean
and standard deviation should be added before submission. Expert-routing maps,
SALF layer-weight maps, parameter count, peak memory, and inference latency
should be reported for A0 and A4.

\subsection{Backbone, Layer, and Capacity Analysis}

Table~\ref{tab:nwpu_val} shows that backbone scaling is non-monotonic. Among the
ConvNeXt variants, Tiny obtains 36.5/94.0, outperforming Small (40.3/114.7), Base
(38.3/122.6), and Large (39.9/124.0). These runs do not isolate backbone capacity
because their decoder widths vary from FPN384 to FPN768, and a matched
fixed-fusion ConvNeXt-Base control is not yet available. Under ConvNeXt-Large,
FPN768 obtains a paired 39.9/124.0, whereas FPN512 gives independently selected
minima of 40.1/128.1; the difference is small and does not establish a general
benefit from wider decoders.

The ViT series improves more consistently up to ViT-L/16: ViT-S/16, ViT-S+/16,
ViT-B/16, and ViT-L/16 obtain 39.3/104.4, 36.5/87.8, 32.2/75.9, and 31.7/72.2,
respectively. ViT-L/16 is the strongest paired result. The ViT-H+/16 entry,
31.0/72.7, combines independently selected minima and therefore does not
establish a better paired operating point. Backbone and decoder width also change
together across these variants, so the trend should be interpreted as a
model-scale observation rather than an isolated backbone ablation.

Layer selection has a substantial but backbone-dependent effect. For ViT-B/16,
replacing blocks $[1,8,10,11]$ with the last four blocks $[8,9,10,11]$ improves
the independently selected minima from 40.5/118.1 to 36.7/87.2. For ViT-L/16
with FPN512, the paired last-four result of 31.7/72.2 is markedly better than the
cross-depth minima of 35.8/115.5. In contrast, under the matched ViT-H+/16
FPN1024 setting, cross-depth blocks $[10,11,30,31]$ improve the paired result
from 33.3/90.1 to 31.6/83.0. These results indicate that useful feature depth
depends on the backbone rather than following one universal selection rule.

Capacity and resolution show similar interactions. Reducing the ViT-B/16 crop and
sliding window from 448 to 224 worsens the independently selected minima from
36.7/87.2 to 39.8/93.6. For ViT-H+/16 cross-depth fusion, FPN1024 yields a paired
31.6/83.0, while FPN512 reaches lower independently selected minima of 31.0/72.7;
because the latter metrics occur at different epochs, they cannot support a
paired width comparison. Overall, neither a larger frozen backbone nor a wider
decoder guarantees lower error. Detailed H+ parameter and training settings are
retained in Tables~\ref{tab:hplus_params} and~\ref{tab:hplus_train}, and per-run
epochs, metrics, and checkpoint paths remain in Table~\ref{tab:record}.

\subsection{Discussion}

The ConvNeXt experiments show that DCA-MoE can consume hierarchical stage
features, but the Small, Tiny, Base, and Large runs also change backbone capacity
and, in some cases, decoder width. They therefore support robustness and scale
analysis rather than a pure component claim. For ViT, the last-four-block
comparison and the cross-depth runs indicate that layer selection materially
affects dense counting. In particular, B7 using $[8,9,10,11]$ outperforms B6
using $[1,8,10,11]$ under the fixed-fusion protocol, while the ViT-L cross-depth
run $[10,11,22,23]$ is weaker than the last-four configuration. The completed
seed-42 A0-A4 matrix provides a fixed-width component comparison: A2 measures
DR-MoE without SALF, A3 measures the full architecture without balancing, and A4
adds the balancing objective. The independently selected minima indicate a clear
gain from SALF and a smaller additional gain from router balancing, while
repeated-seed statistics are still needed to quantify uncertainty.

\section{Conclusion}

This work presents DCA-MoE for crowd counting. SALF learns where to draw
information from across backbone depths, while DR-MoE learns how much local,
mid-range, and large-context processing each position requires. The modules are
optimized jointly with block-density reconstruction, DMCount supervision, and
router balancing while the DINOv3 backbone remains frozen. NWPU-Crowd validation
experiments demonstrate that the design is viable across ConvNeXt and ViT
encoders, with a paired MAE/RMSE of 31.7421/72.2007 for the ViT-L/16
configuration. Under the matched ViT-B/16 FPN384 seed-42 ablation, Full w/o
balance obtains independently selected best MAE/RMSE values of 32.7986/77.6230,
while the complete DCA-MoE obtains a paired 32.2124/75.8733. The result indicates
that the full architecture provides the main improvement and router balancing
offers a smaller additional gain. Paired metrics for A0-A3, repeated-seed
statistics, and cross-dataset evaluation remain necessary before stronger claims
about generality.

\appendix

\section{Experiment Configuration Index}

\begin{table}[t]
  \centering
  \caption{Experiment configuration index.}
  \label{tab:config_index}
  \small
  \begin{tabular}{@{}lp{0.23\textwidth}p{0.40\textwidth}l@{}}
    \toprule
    ID & Purpose & Backbone/layer setting & Loss \\
    \midrule
    B0 & reference implementation & CLIP-EBC official configuration & official \\
    B1 & main DINO baseline & ConvNeXt-Base, all four stages & DMCount \\
    B2 & architecture-independent comparison & ViT-L/16, $[1,17,21,23]$ & DMCount \\
    B3 & layer-selection control & ViT-L/16, $[4,11,17,23]$ & DMCount \\
    B4 & objective control & ConvNeXt-Base & CE + MAE \\
    B5 & objective control & ConvNeXt-Base & CE + MSE \\
    B6 & controlled ViT baseline & ViT-B/16, $[1,8,10,11]$ & DMCount \\
    B7 & last-four-layer control & ViT-B/16, $[8,9,10,11]$ & DMCount \\
    B7-224 & input-resolution control & ViT-B/16, $[8,9,10,11]$, $224\times224$ input & DMCount \\
    B7-MoE & capacity-scaled DCA-MoE & ViT-B/16, $[8,9,10,11]$, \texttt{fpn384}, \texttt{lambda\_route=0.1} & DMCount \\
    C-S-MoE & model-scale DCA-MoE & ConvNeXt-Small, four stages, \texttt{fpn384}, \texttt{lambda\_route=0.1} & DMCount \\
    C-T-MoE & model-scale DCA-MoE & ConvNeXt-Tiny, four stages, \texttt{fpn384}, \texttt{lambda\_route=0.1} & DMCount \\
    C-MoE & capacity-scaled adaptive decoder & ConvNeXt-Base, four stages, \texttt{fpn512}, \texttt{lambda\_route=0.1} & DMCount \\
    C-L-MoE & model-scale adaptive decoder & ConvNeXt-Large, four stages, \texttt{fpn768}, \texttt{lambda\_route=0.1} & DMCount \\
    C-L-MoE-512 & decoder-width control & ConvNeXt-Large, four stages, \texttt{fpn512}, \texttt{lambda\_route=0.1} & DMCount \\
    S-MoE & model-scale adaptive decoder & ViT-S/16, $[8,9,10,11]$, \texttt{fpn192}, \texttt{lambda\_route=0.1} & DMCount \\
    S+-MoE & model-scale adaptive decoder & ViT-S+/16, last four $[8,9,10,11]$, \texttt{fpn192}, \texttt{lambda\_route=0.1} & DMCount \\
    L-MoE & model-scale adaptive decoder & ViT-L/16, last four $[20,21,22,23]$, \texttt{fpn512}, \texttt{lambda\_route=0.1} & DMCount \\
    L-MoE-explicit & repeatability check & ViT-L/16, explicitly selected $[20,21,22,23]$, \texttt{fpn512}, \texttt{lambda\_route=0.1} & DMCount \\
    L-MoE-CD & layer-selection control & ViT-L/16, $[10,11,22,23]$, \texttt{fpn512}, \texttt{lambda\_route=0.1} & DMCount \\
    H+-MoE & model-scale adaptive decoder & ViT-H+/16, last four $[28,29,30,31]$, \texttt{fpn1024}, \texttt{lambda\_route=0.1} & DMCount \\
    H+-MoE-1024-CD & layer-selection control & ViT-H+/16, $[10,11,30,31]$, \texttt{fpn1024}, \texttt{lambda\_route=0.1} & DMCount \\
    H+-MoE-512 & decoder-width control & ViT-H+/16, last four $[28,29,30,31]$, \texttt{fpn512}, \texttt{lambda\_route=0.1} & DMCount \\
    H+-MoE-512-CD & layer-selection control & ViT-H+/16, $[10,11,30,31]$, \texttt{fpn512}, \texttt{lambda\_route=0.1} & DMCount \\
    \bottomrule
  \end{tabular}
\end{table}

The B1 comparison with the paper-reported CLIP-EBC result should be presented as
a reference comparison. A separately reimplemented CLIP-EBC row is required
before claiming a controlled backbone comparison. B2-B5 are supporting
experiments and should not be described as new contributions.

The detailed ViT-H+/16 model and training configurations are retained here for
reproducibility. Parameter counts include convolution and GroupNorm affine
parameters; the anchor-point buffer is not trainable.

\begin{table}[t]
  \centering
  \caption{ViT-H+/16 DCA-MoE parameter breakdown.}
  \label{tab:hplus_params}
  \small
  \begin{tabular}{@{}p{0.22\textwidth}p{0.40\textwidth}r l@{}}
    \toprule
    Component & Configuration & Parameters & Optimization status \\
    \midrule
    DINOv3 ViT-H+/16 backbone & hidden size 1280; 32 blocks; 20 heads; MLP size 5120; patch size 16; 4 register tokens & 840,592,640 & frozen \\
    Lateral projections & four $1\times1$ convolutions, 1280 $\to$ 1024 & 5,246,976 & trainable \\
    Smooth block & $3\times3$ convolution, 1024 channels, GroupNorm(32) & 9,440,256 & trainable \\
    Layer-fusion gate & $1\times1$ convolutions, 4096 $\to$ 1024 $\to$ 4, GroupNorm(32) & 4,200,452 & trainable \\
    Local-detail expert & full $3\times3$ convolution + $1\times1$ convolution, 1024 channels & 10,489,856 & trainable \\
    Mid-range expert & depthwise $3\times3$ dilation-2 + $1\times1$ convolution & 1,061,888 & trainable \\
    Large-context expert & depthwise $7\times7$ + $1\times1$ convolution & 1,102,848 & trainable \\
    Density router & $1\times1$ convolutions, 1024 $\to$ 1024 $\to$ 3, GroupNorm(32) & 1,053,699 & trainable \\
    Classifier & $3\times3$ convolution 1024 $\to$ 1024, then $1\times1$ convolution 1024 $\to$ 5 bins & 9,443,333 & trainable \\
    Trainable decoder total & lateral + smooth + gate + experts + router + classifier & 42,039,308 & trainable \\
    Complete model & frozen backbone + trainable decoder & 882,631,948 & total \\
    \bottomrule
  \end{tabular}
\end{table}

\begin{table}[t]
  \centering
  \caption{ViT-H+/16 DCA-MoE training configuration.}
  \label{tab:hplus_train}
  \small
  \begin{tabular}{@{}p{0.30\textwidth}p{0.64\textwidth}@{}}
    \toprule
    Training item & Value \\
    \midrule
    Dataset & NWPU-Crowd \\
    Input and output resolution & $448\times448$ input/window/stride; output reduction 8 \\
    Count objective & DMCount; \texttt{weight\_count\_loss=1.0}, \texttt{weight\_ot=0.1}, \texttt{weight\_tv=0.01} \\
    Router regularization & \texttt{router\_loss\_weight=0.1}; 3 experts \\
    Optimizer and schedule & AdamW; initial learning rate $1\times10^{-4}$; weight decay $1\times10^{-4}$; 5 warm-up epochs; cosine decay to 0.01 of the initial rate \\
    Mixed precision & BF16 AMP \\
    Distributed training & 8 GPUs; \texttt{world\_size=8}; per-GPU batch size 8; global batch size 64 \\
    Training duration & 2000 epochs; validation starts at epoch 1200 \\
    Random seed & 42 \\
    \bottomrule
  \end{tabular}
\end{table}

\section{Reproducibility Commands}

\subsection{Preprocess NWPU}

\begin{lstlisting}[language=bash]
python preprocess.py --dataset nwpu \
  --src_dir ./raw_data/NWPU-Crowd \
  --dst_dir ./data/nwpu \
  --min_size 448 \
  --max_size 3072
\end{lstlisting}

\subsection{Train the ConvNeXt baseline on one GPU}

\begin{lstlisting}[language=bash]
python train.py \
  --dataset nwpu \
  --data_root ./data \
  --dino_path ./facebook/dinov3-convnext-base-pretrain-lvd1689m \
  --output_dir ./checkpoints \
  --batch_size 8 \
  --epochs 2000 \
  --fpn_channels 128 \
  --count_loss dmcount \
  --eval_start 1200 \
  --sliding_window \
  --amp
\end{lstlisting}

\subsection{Train the capacity-scaled ConvNeXt-Base adaptive-MoE run}

\begin{lstlisting}[language=bash]
setsid torchrun --standalone --nproc_per_node 8 train.py \
  --dataset nwpu \
  --data_root ./data \
  --dino_path ./facebook/dinov3-convnext-base-pretrain-lvd1689m \
  --output_dir ./checkpoints \
  --batch_size 8 \
  --epochs 2000 \
  --count_loss dmcount \
  --eval_start 1200 \
  --sliding_window \
  --amp \
  --router_loss_weight 0.1 > train.log 2>&1
\end{lstlisting}

With the current defaults, this command uses all four DINOv3 ConvNeXt-Base
stages, \texttt{fpn\_channels=512}, $448\times448$ input/window/stride, BF16 AMP,
and the adaptive three-expert decoder. The best validation MAE/RMSE are
38.2541/122.5547 at epoch 1257. At epoch 2000, the validation MAE/RMSE are
44.3295/185.5725 and the router balance loss is 1.020021.

Run directory:

\begin{lstlisting}
checkpoints/nwpu/dinov3_convnext_base_ebc_dmcount_ot0.1_tv0.01_fpn512_adaptive_moe
\end{lstlisting}

\subsection{Train the ConvNeXt-Small DCA-MoE run}

\begin{lstlisting}[language=bash]
torchrun --standalone --nproc_per_node 8 train.py \
  --dataset nwpu \
  --data_root ./data \
  --dino_path ./facebook/dinov3-convnext-small-pretrain-lvd1689m \
  --output_dir ./checkpoints \
  --batch_size 8 \
  --epochs 2000 \
  --count_loss dmcount \
  --eval_start 1200 \
  --sliding_window \
  --amp \
  --router_loss_weight 0.1 > train.log 2>&1
\end{lstlisting}

This command uses all four DINOv3 ConvNeXt-Small stages,
\texttt{fpn\_channels=384}, $448\times448$ input/window/stride, BF16 AMP, SALF,
and DR-MoE. The best validation MAE/RMSE are 40.3218/114.6859 at epoch 1214. At
epoch 2000, the validation MAE/RMSE are 45.8800/175.9689 and the router balance
loss is 1.062345.

Run directory:

\begin{lstlisting}
checkpoints/nwpu/dinov3_dinov3_convnext_small_pretrain_lvd1689m_ebc_dmcount_ot0.1_tv0.01_fpn384_adaptive_moe
\end{lstlisting}

\subsection{Train the ConvNeXt-Tiny DCA-MoE run}

\begin{lstlisting}[language=bash]
setsid torchrun --standalone --nproc_per_node 8 train.py \
  --dataset nwpu \
  --data_root ./data \
  --dino_path ./facebook/dinov3-convnext-tiny-pretrain-lvd1689m \
  --output_dir ./checkpoints \
  --batch_size 8 \
  --epochs 2000 \
  --count_loss dmcount \
  --eval_start 1200 \
  --sliding_window \
  --amp \
  --router_loss_weight 0.1 > train.log 2>&1
\end{lstlisting}

This command uses all four DINOv3 ConvNeXt-Tiny stages,
\texttt{fpn\_channels=384}, $448\times448$ input/window/stride, BF16 AMP, SALF,
and DR-MoE. The best validation MAE/RMSE are 36.5369/94.0043 at epoch 1384. At
epoch 2000, the validation MAE/RMSE are 44.0643/151.2707 and the router balance
loss is 1.040030.

Run directory:

\begin{lstlisting}
checkpoints/nwpu/dinov3_dinov3_convnext_tiny_pretrain_lvd1689m_ebc_dmcount_ot0.1_tv0.01_fpn384_adaptive_moe
\end{lstlisting}

\subsection{Train the capacity-scaled ConvNeXt-Large adaptive-MoE run}

\begin{lstlisting}[language=bash]
setsid torchrun --standalone --nproc_per_node 8 train.py \
  --dataset nwpu \
  --data_root ./data \
  --dino_path ./facebook/dinov3-convnext-large-pretrain-lvd1689m \
  --output_dir ./checkpoints \
  --batch_size 8 \
  --epochs 2000 \
  --count_loss dmcount \
  --eval_start 1200 \
  --sliding_window \
  --amp \
  --router_loss_weight 0.1 > train.log 2>&1
\end{lstlisting}

With the current defaults, this command uses all four DINOv3 ConvNeXt-Large
stages, \texttt{fpn\_channels=768}, $448\times448$ input/window/stride, BF16 AMP,
and the adaptive three-expert decoder. The best validation MAE/RMSE are
39.8941/123.9854 at epoch 1296. At epoch 2000, the validation MAE/RMSE are
46.1796/178.0984 and the router balance loss is 1.085159.

Run directory:

\begin{lstlisting}
checkpoints/nwpu/dinov3_dinov3_convnext_large_pretrain_lvd1689m_ebc_dmcount_ot0.1_tv0.01_fpn768_adaptive_moe
\end{lstlisting}

\subsection{Train the ConvNeXt-Large adaptive-MoE run with 512 decoder channels}

\begin{lstlisting}[language=bash]
setsid torchrun --standalone --nproc_per_node 8 train.py \
  --dataset nwpu \
  --data_root ./data \
  --dino_path ./facebook/dinov3-convnext-large-pretrain-lvd1689m \
  --output_dir ./checkpoints \
  --batch_size 8 \
  --epochs 2000 \
  --fpn_channels 512 \
  --count_loss dmcount \
  --eval_start 1200 \
  --sliding_window \
  --amp \
  --router_loss_weight 0.1 > train.log 2>&1
\end{lstlisting}

The current training script defines the option as \texttt{--fpn\_channels}
(plural). The supplied transcript uses \texttt{--fpn\_channel}, while the
completed run directory and its name confirm the effective \texttt{fpn512}
configuration. This run uses all four DINOv3 ConvNeXt-Large stages,
$448\times448$ input/window/stride, BF16 AMP, and the adaptive three-expert
decoder. The best validation MAE/RMSE are 40.1075/128.1433 at epochs 1536/1225,
respectively. At epoch 2000, the validation MAE/RMSE are 44.5100/159.4389 and the
router balance loss is 1.019843.

Run directory:

\begin{lstlisting}
checkpoints/nwpu/dinov3_dinov3_convnext_large_pretrain_lvd1689m_ebc_dmcount_ot0.1_tv0.01_fpn512_adaptive_moe
\end{lstlisting}

\subsection{Train the ViT-L layer-selection baseline}

\begin{lstlisting}[language=bash]
python train.py \
  --dataset nwpu \
  --data_root ./data \
  --dino_path ./facebook/dinov3-vitl16-pretrain-lvd1689m \
  --output_dir ./checkpoints \
  --batch_size 2 \
  --epochs 2000 \
  --fpn_channels 128 \
  --vit_layers "1,17,21,23" \
  --count_loss dmcount \
  --eval_start 1200 \
  --sliding_window \
  --amp
\end{lstlisting}

\subsection{Train the controlled ViT-B/16 fixed-fusion baseline}

\begin{lstlisting}[language=bash]
setsid torchrun --standalone --nproc_per_node 8 train.py \
  --dataset nwpu \
  --data_root ./data \
  --dino_path ./facebook/dinov3-vitb16-pretrain-lvd1689m \
  --output_dir ./checkpoints \
  --batch_size 8 \
  --epochs 2000 \
  --fpn_channels 128 \
  --vit_layers "1,8,10,11" \
  --count_loss dmcount \
  --eval_start 1200 \
  --sliding_window \
  --amp \
  --decoder_variant baseline > train.log 2>&1
\end{lstlisting}

Run directory:

\begin{lstlisting}
checkpoints/nwpu/dinov3_vitb16_ebc_dmcount_ot0.1_tv0.01_fpn128_baseline_layers1_8_10_11
\end{lstlisting}

\subsection{Train the controlled ViT-B/16 last-four-layer baseline}

\begin{lstlisting}[language=bash]
setsid torchrun --standalone --nproc_per_node 8 train.py \
  --dataset nwpu \
  --data_root ./data \
  --dino_path ./facebook/dinov3-vitb16-pretrain-lvd1689m \
  --output_dir ./checkpoints \
  --batch_size 8 \
  --epochs 2000 \
  --fpn_channels 128 \
  --count_loss dmcount \
  --eval_start 1200 \
  --sliding_window \
  --input_size 448 \
  --window_size 448 \
  --stride 448 \
  --amp \
  --decoder_variant baseline > train.log 2>&1
\end{lstlisting}

Run directory:

\begin{lstlisting}
checkpoints/nwpu/dinov3_vitb16_ebc_dmcount_ot0.1_tv0.01_fpn128_baseline_layers8_9_10_11
\end{lstlisting}

For the capacity-matched A0 baseline used in the component ablation, keep the
same last-four layers and change only the decoder width to 384. Use a clean
output root so that the run cannot resume the historical FPN128 checkpoint:

\begin{lstlisting}[language=bash]
setsid torchrun --standalone --nproc_per_node 8 train.py \
  --dataset nwpu \
  --data_root ./data \
  --dino_path ./facebook/dinov3-vitb16-pretrain-lvd1689m \
  --output_dir ./checkpoints_ablation/A0_baseline_seed42 \
  --batch_size 8 \
  --epochs 2000 \
  --fpn_channels 384 \
  --vit_layers "8,9,10,11" \
  --count_loss dmcount \
  --eval_start 1200 \
  --sliding_window \
  --input_size 448 \
  --window_size 448 \
  --stride 448 \
  --amp \
  --router_loss_weight 0 \
  --decoder_variant baseline > train.log 2>&1
\end{lstlisting}

This A0 run obtains best validation MAE/RMSE of 36.1943/84.6734 at epochs
1586/1229, respectively. At epoch 2000, it reports 38.5958/103.1175. The
checkpoint directory is:

\begin{lstlisting}
checkpoints_ablation/A0_baseline_seed42/nwpu/dinov3_vitb16_ebc_dmcount_ot0.1_tv0.01_fpn384_baseline_layerslast4
\end{lstlisting}

\subsection{Train the matched ViT-B/16 SALF-only ablation}

Use a clean output root so that the run cannot resume an existing baseline or
adaptive-MoE checkpoint:

\begin{lstlisting}[language=bash]
setsid torchrun --standalone --nproc_per_node 8 train.py \
  --dataset nwpu \
  --data_root ./data \
  --dino_path ./facebook/dinov3-vitb16-pretrain-lvd1689m \
  --output_dir ./checkpoints_ablation/A1_salf_only_seed42 \
  --batch_size 8 \
  --epochs 2000 \
  --fpn_channels 384 \
  --count_loss dmcount \
  --eval_start 1200 \
  --sliding_window \
  --input_size 448 \
  --window_size 448 \
  --stride 448 \
  --amp \
  --router_loss_weight 0 \
  --decoder_variant salf_only > train.log 2>&1
\end{lstlisting}

The SALF-only decoder computes spatial softmax weights over the four aligned
backbone features and then applies the shared decoder. It does not instantiate
\texttt{experts} or \texttt{density\_router}, and the training log therefore omits
\texttt{router\_balance\_loss}. The expected checkpoint directory is:

\begin{lstlisting}
checkpoints_ablation/A1_salf_only_seed42/nwpu/dinov3_vitb16_ebc_dmcount_ot0.1_tv0.01_fpn384_salf_only_layerslast4
\end{lstlisting}

\subsection{Train the matched ViT-B/16 MoE-only ablation}

Use a separate output root to prevent automatic resume from another decoder
variant:

\begin{lstlisting}[language=bash]
setsid torchrun --standalone --nproc_per_node 8 train.py \
  --dataset nwpu \
  --data_root ./data \
  --dino_path ./facebook/dinov3-vitb16-pretrain-lvd1689m \
  --output_dir ./checkpoints_ablation/A2_moe_only_seed42 \
  --batch_size 8 \
  --epochs 2000 \
  --fpn_channels 384 \
  --vit_layers "8,9,10,11" \
  --count_loss dmcount \
  --eval_start 1200 \
  --sliding_window \
  --input_size 448 \
  --window_size 448 \
  --stride 448 \
  --amp \
  --router_loss_weight 0.1 \
  --decoder_variant moe_only > train.log 2>&1
\end{lstlisting}

The expected checkpoint directory is:

\begin{lstlisting}
checkpoints_ablation/A2_moe_only_seed42/nwpu/dinov3_vitb16_ebc_dmcount_ot0.1_tv0.01_fpn384_moe_only_layers8_9_10_11
\end{lstlisting}

The supplied single-seed run used the same MoE-only decoder configuration, but
its command output root was \texttt{./checkpoints\_ablation/A1\_salf\_only\_seed42}.
It reached a best validation MAE of 36.1265 at epoch 1478 and a best validation
RMSE of 89.0765 at epoch 1299. At epoch 2000, MAE/RMSE were 39.2971/116.4205.
Because the root directory name does not match the decoder variant, the run is
classified by the explicit \texttt{--decoder\_variant moe\_only} argument, and
future reproductions should use the clean A2 output root shown above. The actual
checkpoint path is:

\begin{lstlisting}
checkpoints_ablation/A1_salf_only_seed42/nwpu/dinov3_vitb16_ebc_dmcount_ot0.1_tv0.01_fpn384_moe_only_layerslast4
\end{lstlisting}

\subsection{Train the matched ViT-B/16 Full w/o balance ablation}

This run retains SALF, all three receptive-field experts, and the spatial router,
while removing only the auxiliary router-balancing term:

\begin{lstlisting}[language=bash]
setsid torchrun --standalone --nproc_per_node 8 train.py \
  --dataset nwpu \
  --data_root ./data \
  --dino_path ./facebook/dinov3-vitb16-pretrain-lvd1689m \
  --output_dir ./checkpoints_ablation/A3_full_wo_balance_seed42 \
  --batch_size 8 \
  --epochs 2000 \
  --fpn_channels 384 \
  --vit_layers "8,9,10,11" \
  --count_loss dmcount \
  --eval_start 1200 \
  --sliding_window \
  --input_size 448 \
  --window_size 448 \
  --stride 448 \
  --amp \
  --seed 42 \
  --router_loss_weight 0 \
  --decoder_variant adaptive_moe > train.log 2>&1
\end{lstlisting}

The seed-42 run reaches a best validation MAE of 32.7986 at epoch 1354 and a best
validation RMSE of 77.6230 at epoch 1377. At epoch 2000, MAE/RMSE are
35.4477/91.5442. The two best values are independently selected minima; the
paired RMSE at epoch 1354 must be recovered from the original
\texttt{train\_log.jsonl} or console log before the final table is submitted. The
checkpoint directory reported by the completed run is:

\begin{lstlisting}
checkpoints_ablation/A3_full_wo_balance_seed42/nwpu/dinov3_vitb16_ebc_dmcount_ot0.1_tv0.01_fpn384_adaptive_moe_layers8_9_10_11
\end{lstlisting}

\subsection{Train the controlled ViT-B/16 baseline with 224-pixel input}

\begin{lstlisting}[language=bash]
setsid torchrun --standalone --nproc_per_node 8 train.py \
  --dataset nwpu \
  --data_root ./data \
  --dino_path ./facebook/dinov3-vitb16-pretrain-lvd1689m \
  --output_dir ./checkpoints \
  --batch_size 8 \
  --epochs 2000 \
  --fpn_channels 128 \
  --vit_layers "8,9,10,11" \
  --count_loss dmcount \
  --eval_start 1200 \
  --sliding_window \
  --amp \
  --decoder_variant baseline \
  --input_size 224 \
  --window_size 224 \
  --stride 224 > train.log 2>&1
\end{lstlisting}

Run directory:

\begin{lstlisting}
checkpoints/nwpu/dinov3_vitb16_ebc_dmcount_ot0.1_tv0.01_fpn128_baseline_layers8_9_10_11
\end{lstlisting}

For an independent resolution comparison, this run must not resume the checkpoint
produced by the 448-pixel B7 experiment. The current run-name format does not
encode input resolution and the training script automatically reads
\texttt{last.pt}; therefore, use a clean output root (for example,
\texttt{./checkpoints\_224}) or otherwise archive the 224-pixel checkpoint
separately before rerunning.

\subsection{Train the capacity-scaled ViT-B/16 adaptive-MoE run}

\begin{lstlisting}[language=bash]
setsid torchrun --standalone --nproc_per_node 8 train.py \
  --dataset nwpu \
  --data_root ./data \
  --dino_path ./facebook/dinov3-vitb16-pretrain-lvd1689m \
  --output_dir ./checkpoints \
  --batch_size 8 \
  --epochs 2000 \
  --vit_layers "8,9,10,11" \
  --fpn_channels 384 \
  --router_loss_weight 0.1 \
  --count_loss dmcount \
  --eval_start 1200 \
  --sliding_window \
  --input_size 448 \
  --window_size 448 \
  --stride 448 \
  --amp \
  --decoder_variant adaptive_moe > train.log 2>&1
\end{lstlisting}

Run directory:

\begin{lstlisting}
checkpoints/nwpu/dinov3_vitb16_ebc_dmcount_ot0.1_tv0.01_fpn384_adaptive_moe_layers8_9_10_11
\end{lstlisting}

\subsection{Train the capacity-scaled ViT-S/16 adaptive-MoE run}

\begin{lstlisting}[language=bash]
setsid torchrun --standalone --nproc_per_node 8 train.py \
  --dataset nwpu \
  --data_root ./data \
  --dino_path ./facebook/dinov3-vits16-pretrain-lvd1689m \
  --output_dir ./checkpoints \
  --batch_size 8 \
  --epochs 2000 \
  --vit_layers "8,9,10,11" \
  --fpn_channels 192 \
  --router_loss_weight 0.1 \
  --count_loss dmcount \
  --eval_start 1200 \
  --sliding_window \
  --input_size 448 \
  --window_size 448 \
  --stride 448 \
  --amp \
  --decoder_variant adaptive_moe > train.log 2>&1
\end{lstlisting}

Run directory:

\begin{lstlisting}
checkpoints/nwpu/dinov3_dinov3_vits16_pretrain_lvd1689m_ebc_dmcount_ot0.1_tv0.01_fpn192_adaptive_moe_layers8_9_10_11
\end{lstlisting}

\subsection{Train the capacity-scaled ViT-S+/16 adaptive-MoE run}

\begin{lstlisting}[language=bash]
setsid torchrun --standalone --nproc_per_node 8 train.py \
  --dataset nwpu \
  --data_root ./data \
  --dino_path ./facebook/dinov3-vits16plus-pretrain-lvd1689m \
  --output_dir ./checkpoints \
  --batch_size 8 \
  --epochs 2000 \
  --count_loss dmcount \
  --eval_start 1200 \
  --sliding_window \
  --amp \
  --router_loss_weight 0.1 > train.log 2>&1
\end{lstlisting}

With the current defaults, this command selects the last four ViT-S+/16 blocks
$[8,9,10,11]$, uses \texttt{fpn\_channels=192}, $448\times448$
input/window/stride, BF16 AMP, and the adaptive three-expert decoder. The best
validation MAE/RMSE are 36.5280/87.8401 at epoch 1488. At epoch 2000, the
validation MAE/RMSE are 40.2621/122.8476 and the router balance loss is 1.016136.

Run directory:

\begin{lstlisting}
checkpoints/nwpu/dinov3_dinov3_vits16plus_pretrain_lvd1689m_ebc_dmcount_ot0.1_tv0.01_fpn192_adaptive_moe_layerslast4
\end{lstlisting}

\subsection{Train the capacity-scaled ViT-H+/16 adaptive-MoE run}

\begin{lstlisting}[language=bash]
setsid torchrun --standalone --nproc_per_node 8 train.py \
  --dataset nwpu \
  --data_root ./data \
  --dino_path ./facebook/dinov3-vith16plus-pretrain-lvd1689m \
  --output_dir ./checkpoints \
  --batch_size 8 \
  --epochs 2000 \
  --count_loss dmcount \
  --eval_start 1200 \
  --sliding_window \
  --amp \
  --router_loss_weight 0.1 > train.log 2>&1
\end{lstlisting}

With the current model defaults, this command selects the last four blocks
$[28,29,30,31]$, uses \texttt{fpn\_channels=1024}, and uses $448\times448$
input/window/stride.

Run directory:

\begin{lstlisting}
checkpoints/nwpu/dinov3_dinov3_vith16plus_pretrain_lvd1689m_ebc_dmcount_ot0.1_tv0.01_fpn1024_adaptive_moe_layerslast4
\end{lstlisting}

\subsection{Train the ViT-L/16 adaptive-MoE run}

\begin{lstlisting}[language=bash]
setsid torchrun --standalone --nproc_per_node 8 train.py \
  --dataset nwpu \
  --data_root ./data \
  --dino_path ./facebook/dinov3-vitl16-pretrain-lvd1689m \
  --output_dir ./checkpoints \
  --batch_size 8 \
  --epochs 2000 \
  --count_loss dmcount \
  --eval_start 1200 \
  --sliding_window \
  --amp \
  --router_loss_weight 0.1 > train.log 2>&1
\end{lstlisting}

With the current defaults, the run uses the last four blocks $[20,21,22,23]$,
\texttt{fpn\_channels=512}, $448\times448$ input/window/stride, BF16 AMP, and the
adaptive three-expert decoder. The best validation MAE/RMSE are 31.7421/72.2007
at epoch 1241. At epoch 2000, the validation MAE/RMSE are 32.9023/81.4774 and the
router balance loss is 1.0183.

Run directory:

\begin{lstlisting}
checkpoints/nwpu/dinov3_vitl16_ebc_dmcount_ot0.1_tv0.01_fpn512_adaptive_moe_layerslast4
\end{lstlisting}

\subsection{Repeat the ViT-L/16 run with explicit last-four layers}

\begin{lstlisting}[language=bash]
setsid torchrun --standalone --nproc_per_node 8 train.py \
  --dataset nwpu \
  --data_root ./data \
  --dino_path ./facebook/dinov3-vitl16-pretrain-lvd1689m \
  --output_dir ./checkpoints \
  --batch_size 8 \
  --epochs 2000 \
  --fpn_channels 512 \
  --vit_layers 20,21,22,23 \
  --count_loss dmcount \
  --eval_start 1200 \
  --sliding_window \
  --amp \
  --router_loss_weight 0.1 > train.log 2>&1
\end{lstlisting}

This repeat run explicitly selects ViT-L/16 blocks $[20,21,22,23]$, while
retaining the adaptive three-expert decoder, \texttt{fpn\_channels=512},
$448\times448$ input/window/stride, DMCount, BF16 AMP, and eight-GPU distributed
training. It reaches a best validation MAE of 31.7511 at epoch 1512 and a best
validation RMSE of 75.2289 at epoch 1241. At epoch 2000, the validation MAE/RMSE
are 32.9587/82.2159 and the router balance loss is 1.020173.

Run directory:

\begin{lstlisting}
checkpoints/nwpu/dinov3_vitl16_ebc_dmcount_ot0.1_tv0.01_fpn512_adaptive_moe_layers20_21_22_23
\end{lstlisting}

\subsection{Train the ViT-L/16 cross-depth layer-selection run}

\begin{lstlisting}[language=bash]
setsid torchrun --standalone --nproc_per_node 8 train.py \
  --dataset nwpu \
  --data_root ./data \
  --dino_path ./facebook/dinov3-vitl16-pretrain-lvd1689m \
  --output_dir ./checkpoints \
  --batch_size 8 \
  --epochs 2000 \
  --fpn_channels 512 \
  --vit_layers 10,11,22,23 \
  --count_loss dmcount \
  --eval_start 1200 \
  --sliding_window \
  --amp \
  --router_loss_weight 0.1 > train.log 2>&1
\end{lstlisting}

The run uses ViT-L/16 blocks $[10,11,22,23]$, \texttt{fpn\_channels=512},
$448\times448$ input/window/stride, BF16 AMP, and the adaptive three-expert
decoder. It reached a best validation MAE/RMSE of 35.8094/115.4743 at epochs
1284/1247, respectively. The last complete validation reported was epoch 1761,
with MAE/RMSE 41.8321/181.7899 and router balance loss 1.0133. Training was ended
before the configured 2000 epochs because no further improvement was expected; no
epoch-2000 validation result is available.

Run directory:

\begin{lstlisting}
checkpoints/nwpu/dinov3_vitl16_ebc_dmcount_ot0.1_tv0.01_fpn512_adaptive_moe_layers10_11_22_23
\end{lstlisting}

\subsection{Train the ViT-H+/16 adaptive-MoE run with 512 decoder channels}

\begin{lstlisting}[language=bash]
setsid torchrun --standalone --nproc_per_node 8 train.py \
  --dataset nwpu \
  --data_root ./data \
  --dino_path ./facebook/dinov3-vith16plus-pretrain-lvd1689m \
  --output_dir ./checkpoints \
  --batch_size 8 \
  --epochs 2000 \
  --fpn_channels 512 \
  --count_loss dmcount \
  --eval_start 1200 \
  --sliding_window \
  --amp \
  --router_loss_weight 0.1 > train.log 2>&1
\end{lstlisting}

With the current defaults, the run uses the last four blocks $[28,29,30,31]$,
$448\times448$ input/window/stride, BF16 AMP, and the adaptive three-expert
decoder.

Run directory:

\begin{lstlisting}
checkpoints/nwpu/dinov3_dinov3_vith16plus_pretrain_lvd1689m_ebc_dmcount_ot0.1_tv0.01_fpn512_adaptive_moe_layerslast4
\end{lstlisting}

\subsection{Train the ViT-H+/16 cross-depth layer-selection run}

\begin{lstlisting}[language=bash]
setsid torchrun --standalone --nproc_per_node 8 train.py \
  --dataset nwpu \
  --data_root ./data \
  --dino_path ./facebook/dinov3-vith16plus-pretrain-lvd1689m \
  --output_dir ./checkpoints \
  --batch_size 8 \
  --epochs 2000 \
  --fpn_channels 512 \
  --vit_layers 10,11,30,31 \
  --count_loss dmcount \
  --eval_start 1200 \
  --sliding_window \
  --amp \
  --router_loss_weight 0.1 > train.log 2>&1
\end{lstlisting}

Run directory:

\begin{lstlisting}
checkpoints/nwpu/dinov3_dinov3_vith16plus_pretrain_lvd1689m_ebc_dmcount_ot0.1_tv0.01_fpn512_adaptive_moe_layers10_11_30_31
\end{lstlisting}

\subsection{Train the ViT-H+/16 cross-depth run with 1024 decoder channels}

\begin{lstlisting}[language=bash]
setsid torchrun --standalone --nproc_per_node 8 train.py \
  --dataset nwpu \
  --data_root ./data \
  --dino_path ./facebook/dinov3-vith16plus-pretrain-lvd1689m \
  --output_dir ./checkpoints \
  --batch_size 8 \
  --epochs 2000 \
  --fpn_channels 1024 \
  --vit_layers 10,11,30,31 \
  --count_loss dmcount \
  --eval_start 1200 \
  --sliding_window \
  --amp \
  --router_loss_weight 0.1 > train.log 2>&1
\end{lstlisting}

The run achieved a best validation MAE/RMSE of 31.6277/83.0485 at epoch 1276. At
epoch 2000, the validation MAE/RMSE were 35.8007/120.3435 and the router balance
loss was 1.0588.

Run directory:

\begin{lstlisting}
checkpoints/nwpu/dinov3_dinov3_vith16plus_pretrain_lvd1689m_ebc_dmcount_ot0.1_tv0.01_fpn1024_adaptive_moe_layers10_11_30_31
\end{lstlisting}

\subsection{Train the Cross-Dataset ViT-B/16 A0 Baseline}

Set \texttt{DATASET} to \texttt{sha}, \texttt{shb}, or \texttt{qnrf}. This
final-report command assumes that the training duration has already been fixed on
an internal validation split. Because \texttt{eval\_start} equals
\texttt{epochs}, the local \texttt{val} directory, which represents the official
test split for these datasets, is evaluated only once at the fixed final epoch.
The per-GPU batch size is 1, giving a global batch size of 8 on eight GPUs.

\begin{lstlisting}[language=bash]
DATASET=sha
setsid torchrun --standalone --nproc_per_node 8 train.py \
  --dataset "$DATASET" \
  --data_root ./data \
  --dino_path ./facebook/dinov3-vitb16-pretrain-lvd1689m \
  --output_dir ./checkpoints_cross_dataset/A0_seed42 \
  --batch_size 1 \
  --epochs 2000 \
  --fpn_channels 384 \
  --vit_layers 8,9,10,11 \
  --decoder_variant baseline \
  --count_loss dmcount \
  --eval_start 2000 \
  --eval_freq 1 \
  --sliding_window \
  --amp \
  --seed 42 > "${DATASET}_A0_seed42.log" 2>&1
\end{lstlisting}

\subsection{Train the Cross-Dataset ViT-B/16 Full DCA-MoE Model}

This command changes only the decoder path and router balancing term relative to
the A0 template. Use separate output roots for seeds 42, 43, and 44 to prevent
automatic checkpoint resumption from mixing runs.

\begin{lstlisting}[language=bash]
DATASET=sha
setsid torchrun --standalone --nproc_per_node 8 train.py \
  --dataset "$DATASET" \
  --data_root ./data \
  --dino_path ./facebook/dinov3-vitb16-pretrain-lvd1689m \
  --output_dir ./checkpoints_cross_dataset/A4_seed42 \
  --batch_size 1 \
  --epochs 2000 \
  --fpn_channels 384 \
  --vit_layers 8,9,10,11 \
  --decoder_variant adaptive_moe \
  --count_loss dmcount \
  --eval_start 2000 \
  --eval_freq 1 \
  --sliding_window \
  --amp \
  --router_loss_weight 0.1 \
  --seed 42 > "${DATASET}_A4_seed42.log" 2>&1
\end{lstlisting}

\subsection{Evaluate a Checkpoint}

\begin{lstlisting}[language=bash]
python evaluate.py \
  --dataset nwpu \
  --data_root ./data \
  --dino_path ./facebook/dinov3-convnext-base-pretrain-lvd1689m \
  --checkpoint ./checkpoints/nwpu/<run_name>/best.pt \
  --sliding_window
\end{lstlisting}

\section{Limitations and Future Work}

\subsection{Current Limitations and Required Work}

This is a method-centered working draft. The current evidence is based mainly on
the NWPU-Crowd validation split. Matched component ablations, paired-checkpoint
reporting, repeated runs, efficiency measurements, additional datasets, and
official test-set evaluation remain required before a CCF-B submission.

The current study has four limitations. First, the formal A0/A4 protocol for
ShanghaiTech Part A/B and UCF-QNRF remains incomplete; one provisional SHA
ViT-B/16 test record is available, but its decoder and run metadata are not yet
confirmed, and no official hidden test result is claimed beyond that supplied
record. Second, many exploratory runs change backbone scale and decoder width
together, so they cannot establish that SALF or DR-MoE alone causes the observed
improvement. Third, MAE and RMSE are often selected independently at different
epochs in the existing logs; the main paper should report paired metrics at one
checkpoint and add repeated-seed statistics. Fourth, the current implementation
routes from fused visual features and uses density supervision, but it does not
provide an explicit density-cue input to the router. The phrase density-routed
must therefore retain this precise meaning.

The required next experiments are paired-checkpoint extraction for A0-A3, seeds
43 and 44 for the matched A0-A4 matrix, routing and SALF visualizations,
parameter and latency measurements, and evaluation on ShanghaiTech Part A/B and
UCF-QNRF. The final version should also compare against recent state-of-the-art
methods under identical data splits and clearly separate paper-reported reference
values from reproduced measurements. Backbone fine-tuning and explicit
density-cue routing are possible extensions, but they should not be presented as
current contributions without new code and retraining.

\subsection{Reference-Result Protocol Note}

The CLIP-EBC settings and benchmark values cited in this draft are taken from
Table~III of \cite{ma2025clipebc}. Its ResNet50 model reports MAE/RMSE values of
54.0/83.2 on SHA, 6.0/10.1 on SHB, 80.5/136.6 on UCF-QNRF, and 38.6/90.3 on NWPU
validation. Its ViT-B/16 model reports 52.5/85.9, 6.6/10.5, 80.3/139.3, and
36.6/81.7, respectively. On the NWPU test split, \cite{ma2025clipebc} reports
61.3/278.4 for ViT-B/16 and 58.2/268.5 for ViT-L/14. These values must remain
labeled as published reference results rather than measurements produced by the
current code.

The paper states a block size of 8, truncation parameter 4, Adam with an initial
learning rate of $10^{-4}$, cosine annealing, a batch size of 8 for all datasets,
and 32 learnable visual prompt tokens prepended to each ViT layer. A future
CLIP-EBC reproduction must record the exact official repository revision because
example commands in the repository may differ from the concise paper protocol.
Reproduced results must not replace or silently overwrite the published-reference
rows.

\subsection{Planned Error Analysis}

Figures~\ref{fig:positive_cases} and~\ref{fig:negative_cases} provide initial
positive and negative qualitative evidence. A systematic error analysis should
extend these visualizations to the following cases:

\begin{enumerate}
  \item sparse scenes, where the count is dominated by a small number of isolated
        people;
  \item dense scenes, where local peaks merge and count errors are amplified;
  \item strong perspective changes, where head scale varies significantly within
        an image;
  \item severe occlusion or background clutter, where semantic features may
        suppress weak local evidence;
  \item very large images evaluated with and without sliding-window inference.
\end{enumerate}

For each case, report the input image, predicted density map, predicted count,
ground-truth count, and point visualization. The final analysis should sample
the evaluation split systematically rather than retaining only visually favorable
examples.

\section{Experimental Record Sheet}

Fill this section after each run rather than relying only on terminal output.
Metric values are rounded to one decimal place; best MAE and RMSE may come from
different epochs unless the best-epoch cell states otherwise.

\begin{table}[t]
  \centering
  \caption{Experimental record sheet.}
  \label{tab:record}
  \resizebox{\textwidth}{!}{%
  \footnotesize
  \begin{tabular}{l l l c l l c c c l r r l}
    \toprule
    Run ID & Date & Backbone & Input size & Layers/stages / decoder & Loss & GPUs & Per-GPU batch & Epochs & Best epoch & Best MAE & Best RMSE & Checkpoint \\
    \midrule
    NWPU-ConvNeXt-S-MoE & 2026-08-10 & ConvNeXt-Small & 448 & four stages; SALF + DR-MoE, fpn384, lambda\_route=0.1 & DMCount & 8 & 8 & 2000 & 1214 (MAE and RMSE) & 40.3 & 114.7 & \texttt{checkpoints/nwpu/dinov3\_dinov3\_convnext\_small\_pretrain\_lvd1689m\_ebc\_dmcount\_ot0.1\_tv0.01\_fpn384\_adaptive\_moe} \\
    NWPU-ConvNeXt-T-MoE & 2026-08-10 & ConvNeXt-Tiny & 448 & four stages; SALF + DR-MoE, fpn384, lambda\_route=0.1 & DMCount & 8 & 8 & 2000 & 1384 (MAE and RMSE) & 36.5 & 94.0 & \texttt{checkpoints/nwpu/dinov3\_dinov3\_convnext\_tiny\_pretrain\_lvd1689m\_ebc\_dmcount\_ot0.1\_tv0.01\_fpn384\_adaptive\_moe} \\
    NWPU-ConvNeXt-L-MoE-512 & 2026-08-09 & ConvNeXt-Large & 448 & four stages; adaptive MoE, fpn512, lambda\_route=0.1 & DMCount & 8 & 8 & 2000 & 1536 (MAE), 1225 (RMSE) & 40.1 & 128.1 & \texttt{checkpoints/nwpu/dinov3\_dinov3\_convnext\_large\_pretrain\_lvd1689m\_ebc\_dmcount\_ot0.1\_tv0.01\_fpn512\_adaptive\_moe} \\
    NWPU-ConvNeXt-L-MoE & 2026-08-08 & ConvNeXt-Large & 448 & four stages; adaptive MoE, fpn768, lambda\_route=0.1 & DMCount & 8 & 8 & 2000 & 1296 (MAE and RMSE) & 39.9 & 124.0 & \texttt{checkpoints/nwpu/dinov3\_dinov3\_convnext\_large\_pretrain\_lvd1689m\_ebc\_dmcount\_ot0.1\_tv0.01\_fpn768\_adaptive\_moe} \\
    NWPU-ConvNeXt-MoE & 2026-08-06 & ConvNeXt-Base & 448 & four stages; adaptive MoE, fpn512, lambda\_route=0.1 & DMCount & 8 & 8 & 2000 & 1257 (MAE and RMSE) & 38.3 & 122.6 & \texttt{checkpoints/nwpu/dinov3\_convnext\_base\_ebc\_dmcount\_ot0.1\_tv0.01\_fpn512\_adaptive\_moe} \\
    NWPU-B6 & 2026-07-24 & ViT-B/16 & 448 & 1,8,10,11 & DMCount & 8 & 8 & 2000 & 1527 (MAE), 1327 (RMSE) & 40.5 & 118.1 & \texttt{checkpoints/nwpu/dinov3\_vitb16\_ebc\_dmcount\_ot0.1\_tv0.01\_baseline\_layers1\_8\_10\_11} \\
    NWPU-B7 & 2026-07-25 & ViT-B/16 & 448 & 8,9,10,11 & DMCount & 8 & 8 & 2000 & 1363 (MAE), 1247 (RMSE) & 36.7 & 87.2 & \texttt{checkpoints/nwpu/dinov3\_vitb16\_ebc\_dmcount\_ot0.1\_tv0.01\_baseline\_layers8\_9\_10\_11} \\
    NWPU-A0 & 2026-08-11 & ViT-B/16 & 448 & 8,9,10,11; fixed sum, single decoder, fpn384 & DMCount & 8 & 8 & 2000 & 1586 (MAE), 1229 (RMSE) & 36.2 & 84.7 & \texttt{checkpoints\_ablation/A0\_baseline\_seed42/nwpu/dinov3\_vitb16\_ebc\_dmcount\_ot0.1\_tv0.01\_fpn384\_baseline\_layerslast4} \\
    NWPU-A1 & 2026-08-11 & ViT-B/16 & 448 & default last four [8,9,10,11]; SALF-only, single decoder, fpn384 & DMCount & 8 & 8 & 2000 & 1461 (MAE), 1565 (RMSE) & 36.0 & 87.6 & \texttt{checkpoints\_ablation/A1\_salf\_only\_seed42/nwpu/dinov3\_vitb16\_ebc\_dmcount\_ot0.1\_tv0.01\_fpn384\_salf\_only\_layerslast4} \\
    NWPU-A2 & 2026-08-12 & ViT-B/16 & 448 & 8,9,10,11; fixed sum, three experts and router, fpn384, balance on & DMCount & 8 & 8 & 2000 & 1478 (MAE), 1299 (RMSE) & 36.1 & 89.1 & \texttt{checkpoints\_ablation/A1\_salf\_only\_seed42/nwpu/dinov3\_vitb16\_ebc\_dmcount\_ot0.1\_tv0.01\_fpn384\_moe\_only\_layerslast4} (root name mismatch; classified by \texttt{moe\_only}) \\
    NWPU-A3 & 2026-08-12 & ViT-B/16 & 448 & 8,9,10,11; SALF + three experts and router, fpn384, balance off & DMCount & 8 & 8 & 2000 & 1354 (MAE), 1377 (RMSE) & 32.8 & 77.6 & \texttt{checkpoints\_ablation/A3\_full\_wo\_balance\_seed42/nwpu/dinov3\_vitb16\_ebc\_dmcount\_ot0.1\_tv0.01\_fpn384\_adaptive\_moe\_layers8\_9\_10\_11} \\
    NWPU-A4 & 2026-08-11 & ViT-B/16 & 448 & default last four [8,9,10,11]; SALF + three experts and router, fpn384, balance on & DMCount & 8 & 8 & 2000 & 1293 (MAE and RMSE) & 32.2 & 75.9 & \texttt{checkpoints/nwpu/dinov3\_vitb16\_ebc\_dmcount\_ot0.1\_tv0.01\_fpn384\_adaptive\_moe\_layers8\_9\_10\_11} \\
    NWPU-B7-224 & 2026-07-25 & ViT-B/16 & 224 & 8,9,10,11 & DMCount & 8 & 8 & 2000 & 1655 (MAE), 1413 (RMSE) & 39.8 & 93.6 & \texttt{checkpoints/nwpu/dinov3\_vitb16\_ebc\_dmcount\_ot0.1\_tv0.01\_baseline\_layers8\_9\_10\_11} \\
    NWPU-B7-MoE & 2026-07-26 & ViT-B/16 & 448 & 8,9,10,11; adaptive MoE, fpn384, lambda\_route=0.1 & DMCount & 8 & 8 & 2000 & 1293 (MAE and RMSE) & 32.2 & 75.9 & \texttt{checkpoints/nwpu/dinov3\_vitb16\_ebc\_dmcount\_ot0.1\_tv0.01\_fpn384\_adaptive\_moe\_layers8\_9\_10\_11} \\
    NWPU-S-MoE & 2026-07-27 & ViT-S/16 & 448 & 8,9,10,11; adaptive MoE, fpn192, lambda\_route=0.1 & DMCount & 8 & 8 & 2000 & 1273 (MAE), 1212 (RMSE) & 39.3 & 104.4 & \texttt{checkpoints/nwpu/dinov3\_dinov3\_vits16\_pretrain\_lvd1689m\_ebc\_dmcount\_ot0.1\_tv0.01\_fpn192\_adaptive\_moe\_layers8\_9\_10\_11} \\
    NWPU-S+-MoE & 2026-08-06 & ViT-S+/16 & 448 & last four [8,9,10,11]; adaptive MoE, fpn192, lambda\_route=0.1 & DMCount & 8 & 8 & 2000 & 1488 (MAE and RMSE) & 36.5 & 87.8 & \texttt{checkpoints/nwpu/dinov3\_dinov3\_vits16plus\_pretrain\_lvd1689m\_ebc\_dmcount\_ot0.1\_tv0.01\_fpn192\_adaptive\_moe\_layerslast4} \\
    NWPU-H+-MoE & 2026-07-28 & ViT-H+/16 & 448 & last four [28,29,30,31]; adaptive MoE, fpn1024, lambda\_route=0.1 & DMCount & 8 & 8 & 2000 & 1589 (MAE and RMSE) & 33.3 & 90.1 & \texttt{checkpoints/nwpu/dinov3\_dinov3\_vith16plus\_pretrain\_lvd1689m\_ebc\_dmcount\_ot0.1\_tv0.01\_fpn1024\_adaptive\_moe\_layerslast4} \\
    NWPU-H+-MoE-1024-CD & 2026-08-03 & ViT-H+/16 & 448 & layers [10,11,30,31]; adaptive MoE, fpn1024, lambda\_route=0.1 & DMCount & 8 & 8 & 2000 & 1276 (MAE and RMSE) & 31.6 & 83.0 & \texttt{checkpoints/nwpu/dinov3\_dinov3\_vith16plus\_pretrain\_lvd1689m\_ebc\_dmcount\_ot0.1\_tv0.01\_fpn1024\_adaptive\_moe\_layers10\_11\_30\_31} \\
    NWPU-H+-MoE-512 & 2026-07-29 & ViT-H+/16 & 448 & last four [28,29,30,31]; adaptive MoE, fpn512, lambda\_route=0.1 & DMCount & 8 & 8 & 2000 & 1244 (MAE), 1309 (RMSE) & 33.7 & 85.4 & \texttt{checkpoints/nwpu/dinov3\_dinov3\_vith16plus\_pretrain\_lvd1689m\_ebc\_dmcount\_ot0.1\_tv0.01\_fpn512\_adaptive\_moe\_layerslast4} \\
    NWPU-H+-MoE-512-CD & 2026-08-02 & ViT-H+/16 & 448 & layers [10,11,30,31]; adaptive MoE, fpn512, lambda\_route=0.1 & DMCount & 8 & 8 & 2000 & 1413 (MAE), 1278 (RMSE) & 31.0 & 72.7 & \texttt{checkpoints/nwpu/dinov3\_dinov3\_vith16plus\_pretrain\_lvd1689m\_ebc\_dmcount\_ot0.1\_tv0.01\_fpn512\_adaptive\_moe\_layers10\_11\_30\_31} \\
    NWPU-L-MoE & 2026-08-04 & ViT-L/16 & 448 & last four [20,21,22,23]; adaptive MoE, fpn512, lambda\_route=0.1 & DMCount & 8 & 8 & 2000 & 1241 (MAE and RMSE) & 31.7 & 72.2 & \texttt{checkpoints/nwpu/dinov3\_vitl16\_ebc\_dmcount\_ot0.1\_tv0.01\_fpn512\_adaptive\_moe\_layerslast4} \\
    NWPU-L-MoE-explicit & 2026-08-05 & ViT-L/16 & 448 & explicit layers [20,21,22,23]; adaptive MoE, fpn512, lambda\_route=0.1 & DMCount & 8 & 8 & 2000 & 1512 (MAE), 1241 (RMSE) & 31.8 & 75.2 & \texttt{checkpoints/nwpu/dinov3\_vitl16\_ebc\_dmcount\_ot0.1\_tv0.01\_fpn512\_adaptive\_moe\_layers20\_21\_22\_23} \\
    NWPU-L-MoE-CD & 2026-08-04 & ViT-L/16 & 448 & layers [10,11,22,23]; adaptive MoE, fpn512, lambda\_route=0.1 & DMCount & 8 & 8 & 1761 (stopped; 2000 configured) & 1284 (MAE), 1247 (RMSE) & 35.8 & 115.5 & \texttt{checkpoints/nwpu/dinov3\_vitl16\_ebc\_dmcount\_ot0.1\_tv0.01\_fpn512\_adaptive\_moe\_layers10\_11\_22\_23} \\
    SHA-ViT-B-provisional & 2026-08-13 & ViT-B/16 & not reported & decoder, layers, and FPN not reported & not reported & not reported & not reported & not reported & 1438 (MAE and RMSE) & 55.4 & 95.0 & not reported; user-provided SHA test record \\
    SHB-ViT-B-provisional & 2026-08-13 & ViT-B/16 & not reported & decoder, layers, and FPN not reported & not reported & not reported & not reported & not reported & 1971 (MAE), 1923 (RMSE) & 6.9 & 11.4 & not reported; current paired result 8.1/13.3 at an unreported epoch \\
    \bottomrule
  \end{tabular}}
\end{table}

\end{document}